\documentclass[11pt]{article}

\usepackage[final]{acl}

\usepackage{times}
\usepackage{latexsym}

\usepackage[T1]{fontenc}

\usepackage[utf8]{inputenc}

\usepackage{microtype}

\usepackage{inconsolata}

\usepackage{graphicx}
\usepackage{array}
\usepackage{subcaption}
\usepackage{amsfonts}
\usepackage{booktabs}
\usepackage{amsmath}
\usepackage{tabularx}
\usepackage{multirow}
\usepackage{arydshln}
\usepackage{bm}
\usepackage[normalem]{ulem}
\usepackage[table]{xcolor}

\definecolor{delcolor}{RGB}{220, 70, 70}    
\definecolor{paracolor}{RGB}{70, 140, 70}   
\definecolor{structcolor}{RGB}{70, 100, 180} 

\title{Edit Policy-Aligned Text Simplification}

\author{
  \textbf{Xuanxin Wu}\textsuperscript{1},
  \textbf{Yuki Arase}\textsuperscript{2},
  \textbf{Masaaki Nagata}\textsuperscript{3}
\\
 \textsuperscript{1}The University of Osaka,
 \textsuperscript{2}Institute of Science Tokyo,
 \textsuperscript{3}NTT Inc.
\\
    \texttt{xuanxin.wu@ist.osaka-u.ac.jp},
    \texttt{arase@c.titech.ac.jp}
    \\
    \texttt{masaaki.nagata@ntt.com}
}

\begin{document}
\maketitle
\begin{abstract}
Text simplification aims to modify documents and sentences to make them easier to read and understand while preserving their meanings. Different applications require distinct simplification policies, e.g., lexical-oriented simplification and overall rewriting while trading off detail for simplicity. However, achieving such policy-driven control remains an open challenge. In this work, we introduce a simple yet powerful approach that leverages Large Language Model-as-a-Judge (LLM-Judge) to automatically construct policy-aligned training data, removing the need for costly human annotation or parallel corpora. Our method enables the construction of simplification models that adapt to diverse simplification policies at both the sentence and document levels. Remarkably, small-scale open-source LLMs match or surpass GPT-5, as verified by both automatic metrics and edit analysis. The consistent improvements across model families and sizes demonstrate the robustness of our approach.\footnote{Our code and data will be available at \url{https://github.com/WuXuanxin/edit_policy_simp}.}
\end{abstract}

\section{Introduction}
Text simplification benefits users with reading difficulties, such as second-language (L2) learners and people with reading impairments (e.g., dyslexia)~\citep{alva-manchego-etal-2020-data}.
It involves a series of edits, and document- and sentence-level simplification naturally require different editing strategies. 
Sentence-level simplification policy can vary from lexical paraphrasing, deletion, and sentence splitting, while document-level simplification further needs content reordering, discourse-level rewriting, and elaboration on uncommon topics~\citep{xu-etal-2015-problems, wilkens-etal-2020-coreference, vasquez-rodriguez-etal-2023-document, sun-etal-2021-document, laban-etal-2023-swipe}.
The preferred edit policy, i.e., permissible or appropriate edits in given texts, varies significantly depending on the target audience and applications.
In this study, we focus on simplification for L2 education and learning, which has been one of the primary applications of simplification techniques. In this domain, adaptation to a specific edit policy is crucial to achieve effective learning and satisfy teachers' educational goals. However, such adaptation has been out of scope in previous studies.

Recent studies show that large language models (LLMs) such as OpenAI's ChatGPT~\citep{openai2023gpt4} achieve superior performance than previous fine-tuned sequence-to-sequence models through prompt engineering alone, and often generate a mixture of diverse edit types~\citep{kew-etal-2023-bless, heineman-etal-2023-dancing}. 
However, the use of closed-source LLMs in language education is constrained by limited transparency and controllability. Small-scale open-source LLMs present a more feasible option, yet adapting them with policy-driven simplification remains challenging. Key obstacles include: (1) the intrinsic limitations of LLMs, particularly smaller models, which are strong in overall quality but insensitive in following specific edit policies~\citep{barayan-etal-2025-analysing}; and (2) the scarcity of policy-specific parallel simplification corpora. Manual construction of such a parallel corpus is very expensive. No existing studies provide an efficient way, in terms of both data and computational demands, for building simplification models adapted to predefined edit policies. 

Reinforcement learning from human feedback (RLHF)~\citep{ouyang2022traininglanguagemodelsfollow} has proven effective for aligning LLMs with human values. RLHF leverages human preference data rather than parallel corpora. However, collecting human preferences at scale is still costly. Alternatively, LLM-Judge can provide scalable and explainable feedback~\citep{kocmi-federmann-2023-gemba, song2024-automatedessay, niu2024judgerankleveraginglargelanguage}. Building on this, reinforcement learning from AI feedback (RLAIF)~\citep{bai2022constitutionalaiharmlessnessai} appears promising to replace human preference with preferences generated by off-the-shelf LLMs~\citep{tunstall2023zephyrdirectdistillationlm, cao2024enhancing, lee2024RLAIF}. 

In this work, we introduce a framework for policy-aligned simplification that requires neither parallel corpora nor human supervision, while remaining computationally efficient with smaller LLMs. 
We design four edit policies for sentence- and document-level, spanning lexical level, overall rewriting, and elaboration. Our approach guides LLM-Judge to generate high-quality preference data aligned with the edit policies. These data are then used to fine-tune open-source models, including Phi-3-mini-3.8B~\citep{abdin2024phi3technicalreporthighly}, Qwen2.5-7B~\citep{qwen2025qwen25technicalreport}, Llama3.1-8B~\citep{grattafiori2024llama3herdmodels}, and Qwen2.5-14B~\citep{qwen2025qwen25technicalreport}, via light-weight preference optimization algorithms~\citep{xu2024contrastive, xu2025xalma}. Our approach substantially improves the policy alignment capabilities of small-scale LLMs, 
enabling them to match or even surpass GPT-5~\footnote{We used \texttt{gpt-5-2025-08-07} through the OpenAI API.}, as verified by both automatic simplification metrics and edit-type analysis.

\section{Related Work}
The standard approach in previous studies has been supervised training with complex-simple parallel corpora. 
While remarkable progress has been achieved in simplification quality, adaptation to various policies has been out of the scope for both sentence and document simplification. 
Different from these studies, our method does not depend on any parallel corpora and achieves flexible adaptation to various edit-policies. 

\subsection{Sentence Simplification}

\citet{martin-etal-2020-controllable} showed that simple surface-level attributes such as sentence length or lexical difficulty can be controlled through prepended control tokens to the input. The control of difficulty levels has also been explored, which aims to simplify sentences to be appropriate for the target audience of the specific proficiency levels~\citep{scarton-specia-2018-learning, horiguchi-etal-2024-evaluation, li-etal-2025-aligning}. 
With the advent of LLMs, prompt-based techniques have largely surpassed earlier sequence-to-sequence model-based methods in overall simplification quality~\citep{kew-etal-2023-bless, wu2025indepthevaluationlargelanguage}. 
However, prompting offers limited sensitivity to edit policy adaptation, particularly with smaller LLMs~\citep{barayan-etal-2025-analysing}. Recent studies aim to improve edit policy alignment via fine-tuning, however, the policy-specific training signals still largely depend on human supervision~\citep{gao-etal-2026-evaluating}.

\subsection{Document Simplification}
Document simplification has been constrained by the scarcity, quality, and diversity of copora. Existing resources are largely based on Simple English Wikipedia (SEW)\footnote{\url{https://simple.wikipedia.org/wiki/Main_Page}} and Newsela~\citep{alva-manchego-etal-2020-data}, but SEW-based data often requires substantial alignment and can be noisy~\citep{xu-etal-2015-problems,laban-etal-2023-swipe}. Earlier non-LLM approaches trained on these corpora~\citep{blinova-etal-2023-simsum,cripwell-etal-2023-context,vasquez-rodriguez-etal-2023-document} show limited performance and generalization~\citep{bakker-kamps-2024-beyond}. The task is further complicated by the need for coordinated edits across sentences to preserve coherence and meaning~\citep{wilkens-etal-2020-coreference,vasquez-rodriguez-etal-2023-document,laban-etal-2023-swipe}. Recent work shows that LLM-based prompting outperforms earlier non-LLM systems for document simplification~\citep{fang-etal-2025-collaborative,qiang2025redefiningsimplicity}, and LLM outputs can even surpass existing human-written references~\citep{qiang2025redefiningsimplicity}. However, it is also reported that direct prompting often leads to information loss and deletion-heavy behavior~\citep{nandiraju2025automated,fang2026progressive}. 


\section{Edit-Policy Aligned Simplification}
\subsection{Problem Definition}
We train a simplification model using a decoder-only language model $\pi_{\theta}$ parameterized by $\theta$.
Let $\mathcal{P}$ denote a set of simplification policies and $\mathcal{X}$ be a finite set of source texts, either sentences or documents. For $x \in \mathcal{X}$, let $y^{\star}(x,p)$ be the optimal simplification under a policy $p \in \mathcal{P}$.
Our goal is to learn $\pi_{\theta}$ such that
\begin{equation}
\max_{\theta}\;
\mathbb{E}_{x \sim \mathcal{X}}
\big[
\log \pi_{\theta}\!\big(y^{\star}(x,p)\mid x\big)
\big].
\label{eq:ideal-objective}
\end{equation}

Because $y^{\star}(x,p)$ is latent, prior work approximated it relying on human-written simplifications and optimized $\pi_{\theta}$ through supervised fine-tuning~\citep{scarton-specia-2018-learning, martin-etal-2020-controllable, cripwell-etal-2023-document}. In contrast, we build a preference dataset by  LLMs, and optimize $\pi_{\theta}$ using preference optimization.

\subsection{Edit Policies}
In this study, we focus on L2 education and design edit policies informed by research on second language acquisition and language education. All policies are intentionally differentiated by their edit priorities rather than by mutually exclusive edit boundaries, as different edit operations are often interdependent and strict separation may compromise overall text quality.

\subsubsection{Sentence Level Policy}
In L2 education, sentence-level simplification is desired because teachers can easily identify where and how a text has been modified. 
Previous work in both NLP and language education research has shown that the desired type and degree of simplification edits change depending on learner proficiency and readability levels~\citep{agrawal-etal-2021-non, zhong2020discourse}. Specifically, low- to intermediate-level learners benefit from a combination of lexical paraphrasing, structural modifications, and selective deletions to reduce cognitive load.  In contrast, advanced learners benefit from lexical paraphrasing, which supports vocabulary acquisition~\citep{chen2019phrasal}, but they gain comparatively less from added cohesion or deletion~\citep{hosoda2016interplay, zhong2020discourse}. 

Motivated by these findings, we introduce two distinct sentence-level edit policies. As illustrated in Table~\ref{tab:example_fancy1}, \textbf{Sent-Overall} simplification often combines lexical paraphrasing and structural modifications, and deletions to improve readability for intermediate-level language learners. In contrast, \textbf{Sent-Lexical}~\citep{paetzold-specia-2016-understanding, li-etal-2025-aligning} adheres to the original sentence closely while supporting more efficient vocabulary acquisition for advanced learners. 

\begin{table}[t]
\centering
\small
\setlength{\extrarowheight}{2pt}
\resizebox{0.9\columnwidth}{!}{ 
\begin{tabular}{@{}>{\raggedright}p{0.15\linewidth} p{0.8\linewidth}@{}}
\toprule
\textbf{Source} & 
Johan was also the original bassist of the Swedish power metal band HammerFall, but quit before the band ever released a studio album. \\
\midrule
\textbf{Sent-Overall} & 
Johan was \sout{\textcolor{delcolor}{also}} the original bassist \textcolor{structcolor}{for HammerFall, a Swedish power metal band. He} \textcolor{paracolor}{left before they released their first} \sout{\textcolor{delcolor}{studio}} album.
\\
\midrule
\textbf{Sent-Lexical} & 
Johan was also the original bassist of the Swedish power metal band HammerFall, but \textcolor{paracolor}{he left} before \textcolor{paracolor}{they} released \textcolor{paracolor}{any studio albums}. \\
\bottomrule
\end{tabular}
}
\caption{
    Sentence simplifications by our model under two edit policies (Phi-3-mini-3.8B~\citep{abdin2024phi3technicalreporthighly}). We highlight the main simplification edits in each part of the sentence using different colors. 
    \textcolor{paracolor}{Green: Paraphrasing} \quad 
    \textcolor{delcolor}{Red: Delete} \quad 
    \textcolor{structcolor}{Blue: Structural Change}
}
\label{tab:example_fancy1}
\end{table}

\begin{figure}[t]
  \centering
  \includegraphics[width=0.9\columnwidth]{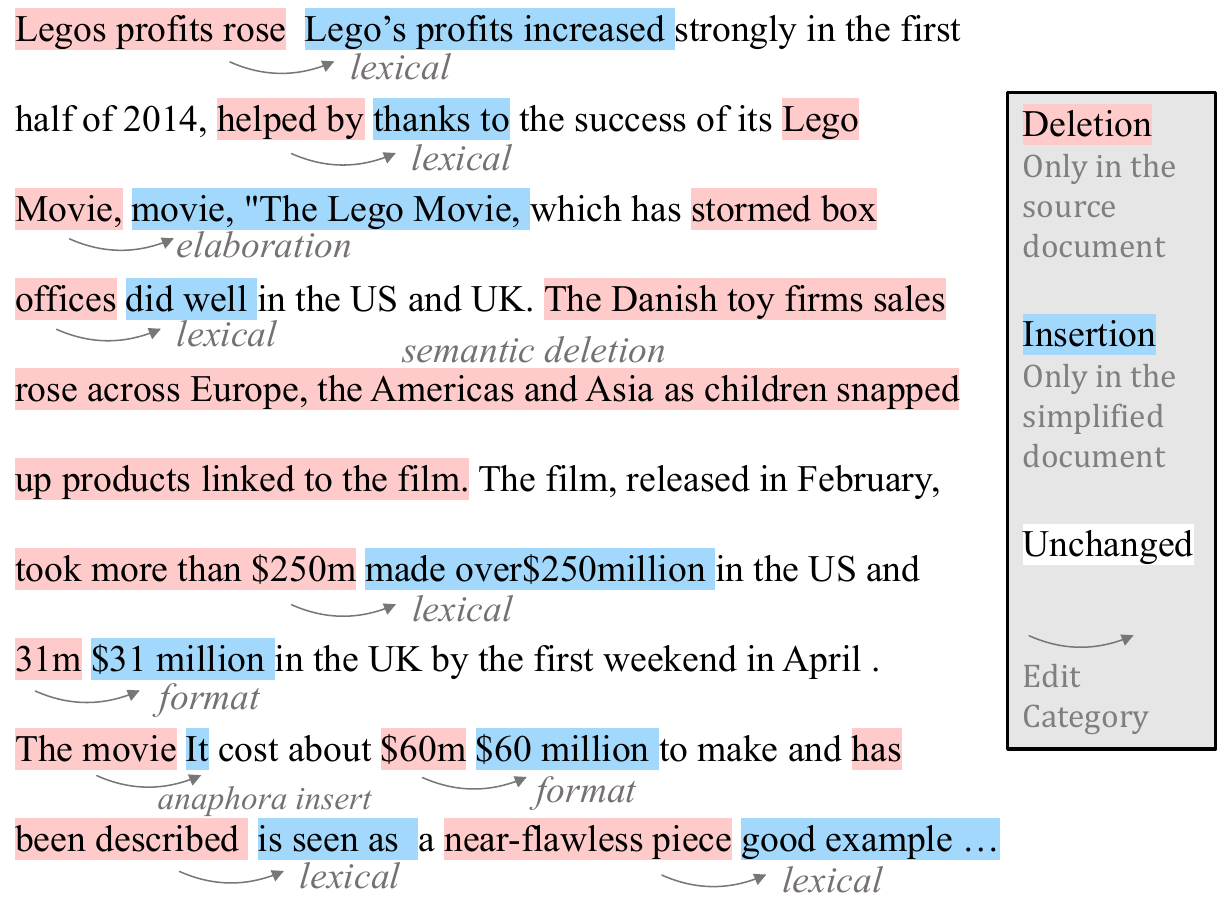}
    \caption{Document simplifications by our model under the Doc-Lexical policy (Llama3.1-8B~\citep{grattafiori2024llama3herdmodels}). Following \citet{laban-etal-2023-swipe}, we visualize the edits applied to the source document.}
  \label{fig:exp_ose}
\end{figure}

\subsubsection{Document Level Policy}
Prior work shows that lexical support in context improves comprehension and vocabulary learning~\citep{chen2019phrasal, Yanagisawa_Webb_Uchihara_2020}, while document comprehension also depends on discourse-level factors~\citep{Horiba_1996}. This suggests that document-level simplification for L2 learners should combine discourse-level rewriting with lexical support, with greater emphasis on lexical accessibility. 
Furthermore, as an orthogonal approach, elaboration is effective to complement the necessary knowledge to understand a content. 
Elaborative edits, i.e., adding clarification and background information, can help reduce readers' cognitive load and improve reading comprehension~\citep{oh2001twotype, srikanth-li-2021-elaborative}. 

Motivated by these findings, we introduce two document-level edit policies. \textbf{Doc-Lexical} prioritizes lexical support (see Figure~\ref{fig:exp_ose}), whereas \textbf{Doc-Elaboration} places greater focus on elaboration.

\subsection{Proposed Method}
\begin{figure*}[t]
  \centering
  \includegraphics[width=0.8\linewidth]{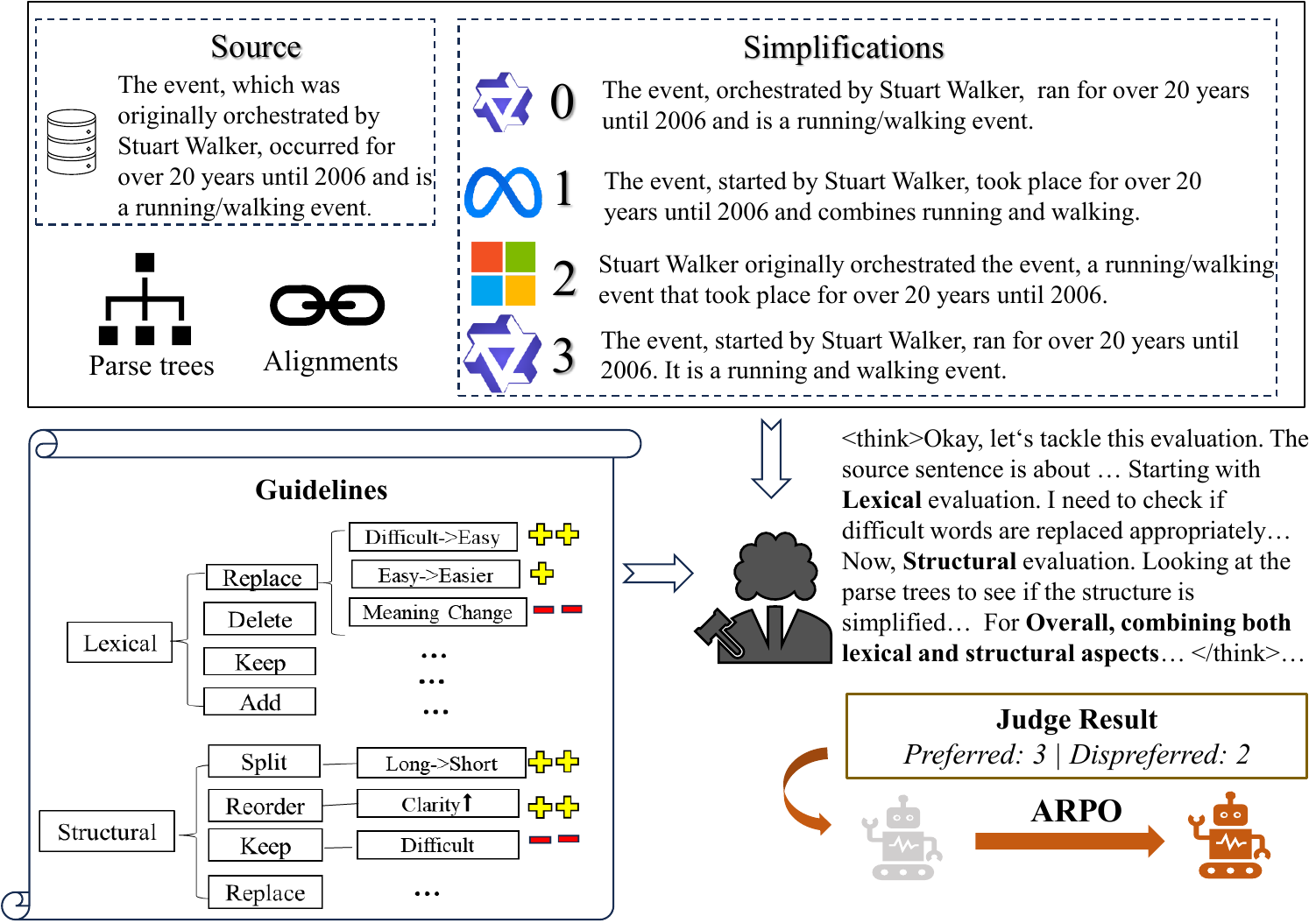}
  \caption{Overview of our sentence-level simplification; the same framework can be applied to the document level.}
  \label{fig:flowchart}
\end{figure*}

Figure~\ref{fig:flowchart} illustrates the overview of our method, which consists of three steps as described below. 

\subsubsection{Step 1: Create Candidate Pool}
We begin with a collection of $N$ source texts $\mathcal{X} = \{x_1, x_2, \ldots, x_N\}$, and use LLMs to generate candidate simplifications. Diversity is crucial to effectively distinguish preferred outputs that align with the target policy from those that deviate from it. Previous studies have shown that models of different families and sizes exhibit distinct performance and editing behavior~\citep{heineman-etal-2023-dancing, kew-etal-2023-bless, wu2025indepthevaluationlargelanguage}. Motivated by this, we construct the candidate pool using a set of $K$ LLMs, $\mathcal{M} = {M_1, \ldots, M_K}$, varying from different families and sizes. For each source text $x_i \in \mathcal{X}$ and policy $p$, every model $M_k \in \mathcal{M}$ generates one candidate simplification $y_{i,k}$, yielding a total of $K$ candidates per source text. The simplification pool for each $x_i$ is defined as:
\begin{equation}
    \mathcal{C}(x_i) = \{y_{i,k} \mid y_{i,k} \sim M_k(x_i, s_p)\},
\label{eq:candidate-pool}
\end{equation}
where $s_p$ is the natural-language instruction that describes policy $p$.

\subsubsection{Step 2: LLM-Judge} 
\label{sec:step2judge}
We employ a reasoning LLM as the judge, selected for its strong performance on complex judge tasks~\citep{huang2026reasoningmodelsuperiorllmjudge}. The model follows carefully designed guidelines for policies. The corresponding prompts are provided in Appendix~\ref{app:prompts_judge}.

\paragraph{Sentence Level Judgement}
The Sent-Lexical policy encourages (a) minimal edits to preserve sentence structure and (b) replacing complex words with simpler alternatives. The Sent-Overall policy encourages broader edits at both lexical and structural levels to enhance simplicity. These nuances can be included in prompts, allowing LLMs to follow them with ease. The model is guided by carefully designed principles that specify edit types, their effects, and associated rewards or penalties.
\begin{itemize}
\item \textbf{Lexical Principles:} We define four edit operations: \textit{replace}, \textit{delete}, \textit{keep}, and \textit{add}. E.g., simplifying a complex word receives a high reward, while replacing an already simple word yields only a moderate reward. 
\item \textbf{Structural Principles:} Similarly, we define four edit operations: \textit{split}, \textit{reorder}, \textit{keep}, and \textit{replace}. Structural transformations are rewarded if they improve readability or conciseness without changing meaning. Unnecessary or unhelpful modifications are penalized.  
\end{itemize}
To support these judgments, we provide the lexical judge with word alignment between the source and simplifications derived using OTAlign~\citep{arase-etal-2023-unbalanced} and the structural judge with syntactic parse trees for each sentence extracted using Qwen3-32B~\citep{yang2025qwen3technicalreport}, respectively. This evidence-guided design improves judge transparency and generally enhances downstream performance. See Appendix~\ref{app:abl_evidence_judge} for details.

\paragraph{Document Level Judgement}
Both of the document-level policies, Doc-Lexical and Doc-Elaboration, involve a mixture of edit types, but differ in the edits they prioritize. 
Assuming that it is difficult for users to verbalize the appropriate compositions of edit types, we introduce an \emph{anchor} example for each policy. 
The use of behavioral anchors has a long history in human evaluation and rating-scale design. Smith and Kendall~\citep{smith-kendall-1963-retranslation} introduced a procedure for constructing rating scales anchored by concrete examples of expected behavior, which became foundational to Behaviorally Anchored Rating Scales (BARS). The underlying idea is to make abstract evaluation criteria more concrete through representative examples. Inspired by this idea, we use an anchor example to concretely instantiate the editing strategy associated with each policy for our LLM-Judge.
Specifically, the anchor illustrates both (a) which edit categories are typically applied and (b) how frequently and strongly they are used under the target policy. To support the decision, we provide edit information for the anchor using a document-level edit classifier~\citep{laban-etal-2023-swipe}. Judges are then required to compare candidate simplifications with the anchor and assess how well their editing strategies align with the target policy, while also considering document-level qualities such as coherence and coreference.

\paragraph{Preference data} 
For each source text $x_i \in \mathcal{X}$, the LLM-judge selects a preferred candidate $y_{w}^{(i)}$ and a dispreferred candidate $y_{l}^{(i)}$ from the candidate pool $\mathcal{C}(x_i)$ according to our guidelines $\mathcal{G}$:
\begin{equation}
(y_{w}^{(i)},\, y_{l}^{(i)}) = J(x_i, \mathcal{C}(x_i,p), \mathcal{G}).
\label{eq:judge-selection}
\end{equation}
This procedure yields a preference dataset $\mathcal{D} = {(x^{(i)}, y_w^{(i)}, y_l^{(i)})}_{i=1}^{N}$ for each policy $p$.

\subsubsection{Step 3: Preference Optimization} 
Preference optimization has emerged as a powerful post-training paradigm for aligning LLMs with human preferences, typically formatted as \texttt{\{input, preferred output, dispreferred output\}}. 
We adopt Adaptive Rejection Preference Optimization (ARPO)~\citep{xu2025xalma} to optimize $\pi_{\theta}$ with our preference dataset $\mathcal{D}$. ARPO is a variant of memory-efficient Contrastive Preference Optimization (CPO)~\citep{xu2024contrastive} designed to mitigate its tendency to overly penalize dispreferred responses that are only marginally worse than preferred ones. We integrate the Simple Preference Optimization (SimPO)~\citep{meng2024simpo} loss into ARPO to improve the performance.\footnote{\url{https://github.com/fe1ixxu/CPO_SIMPO}} See Appendices~\ref{app:preliminary_po}~and~\ref{app:exp_set} for detailed preliminaries and experimental settings.

\section{Evaluation}
\subsection{Preference Data Construction}
\paragraph{Candidate Generation}
We construct the preference optimization dataset using \emph{only} the source side texts from existing simplification corpora. For sentence-level simplification, we use CoEdit~\citep{raheja-etal-2023-coedit}, which aggregates multiple existing simplification parallel corpora. For document-level simplification, we aggregate source documents of Newsela~\citep{xu-etal-2015-problems}\footnote{We only used documents from the highest reading level. Newsela can be requested for academic use at \url{https://newsela.com/legal/data}.} and SWiPE~\cite{laban-etal-2023-swipe}.
For each source text, we collected outputs from four instruction-tuned LLMs: Qwen2.5-7B, Llama3.1-8B, Phi4-14B, and Qwen3-32B. Prompts used to generate simplifications are described in Appendix~\ref{app:prompts_simp}.
To ensure meaningful simplification, we apply heuristic filtering, such as removing very short source texts that leave little room for editing.

\paragraph{LLM-Judge}
We selected Qwen3-32B as the LLM judge based on its strong judging performance in our preliminary experiments. Although Qwen3-32B also serves as a candidate generation model, we did not observe a consistent tendency for it to prefer its own generations. In the sentence-level evaluation, the Qwen3-32B judge in fact prefers outputs from Phi4-14B more frequently. Appendix~\ref{sec:circulate_qwen32b} provides detailed results and discussion.
To mitigate potential positional bias in LLM-based evaluation \citep{wang-etal-2024-large-language-models-fair}, we randomize the presentation order of generated candidates before passing them to the judge model. Consequently, for each policy we obtain a preference dataset consisting of triplets of the form \texttt{\{source, preferred simplification, dispreferred simplification\}}.

\paragraph{Constructed Data}
The sentence-level dataset contains $8$K triplets, split into $7$K for training and $1$K for development. For each document-level policy, we obtained $4$K triplets, split into $3.5$K training instances and $500$ development instances.

\subsection{Evaluation Datasets and Metrics}
\label{sec:eval_data}
\paragraph{Sentence-level Datasets and Metrics} 
For \textbf{Sent-Lexical} policy, we used the \textbf{Turk} test set~\citep{xu-etal-2016-optimizing}, containing $359$ source sentences paired with $8$ crowdworker-written simplification references, constructed specifically for lexical-based simplification. We accordingly use \textbf{SARI}~\citep{xu-etal-2015-problems} as the evaluation metric because it explicitly assesses lexical edit (keep, add, delete) quality by comparing system outputs with both references and sources. For \textbf{Sent-Overall} policy, we used the \textbf{ASSET} test set~\citep{alva-manchego-etal-2020-asset}. While ASSET shares the same source sentences as Turk, it differs in its edit policy: more diverse edits, i.e., paraphrasing, deletion, and sentence splitting. Each source sentence pairs with $10$ crowdworker-written references. To capture this broader range of edits, we evaluated with \textbf{LENS}~\citep{maddela-etal-2023-lens}. LENS is trained directly on human ratings of overall simplicity quality and has shown strong correlation with overall simplicity quality~\citep{huang-kochmar-2024-referee, wu2025indepthevaluationlargelanguage}.

\paragraph{Document-Level Datasets} 
For \textbf{Doc-Lexical}, we use \textbf{OneStopEnglish (OSE)}~\citep{vajjala-lucic-2018-onestopenglish}, an expert-written document simplification dataset created for L2 learners. OSE contains $189$ source documents, each paired with intermediate- and elementary-level simplifications. For \textbf{Doc-Elaboration}, we sample $200$ complex--simple document pairs (not used in the preference data) from expert-written \textbf{Newsela}~\citep{xu-etal-2015-problems} (size: $1.9k$). Newsela includes simplifications at multiple reading levels. 
We use the elementary-level OSE and the middle-level Newsela as references, as we found that they consistently require lexical and elaboration edits while better balance between meaning preservation and simplicity.

\paragraph{Document-Level Metrics} 
We use \textbf{Agg-LENS} \citep{maddela-alva-manchego-2025-adapting}, a document simplification metric that shows higher correlation with human judgments than earlier metrics, e.g., D-SARI~\citep{sun-etal-2021-document}. Agg-LENS adapts the sentence-level LENS metric to the document level through graph-based alignment. We also report \textbf{Agg-BERT} from the same study, which shows high consistency for assessing coherence.
Furthermore, because Agg-LENS is not explicitly designed to capture the edit policies studied in this work, we additionally report \textbf{edit distribution}. To do so, we use an edit-category classifier\footnote{\url{https://github.com/salesforce/simplification}} \cite{laban-etal-2023-swipe}, which was trained on a large collection of human-annotated document-level edits. We measure the percentage of \emph{lexical} edits for Doc-Lexical and \emph{elaboration} edits for Doc-Elaboration. We also report \emph{deletion}, as prior work has shown that LLMs tend to over-delete content~\citep{nandiraju2025automated, fang2026progressive}. These distributions are then compared with those of the reference simplifications. Implementation details are in Appendix~\ref{app:edit_dist}.

\subsection{Comparison}
\label{sec:baselines}

We apply preference optimization to four open-source instruction-tuned LLMs for sentence-level policies, covering different model families and scales: Phi-3-mini-3.8B (\texttt{Phi3.8B}), Qwen2.5-7B (\texttt{Qwen7B}), Llama3.1-8B (\texttt{Llama8B}), and Qwen2.5-14B (\texttt{Qwen14B}). For document-level, we use Qwen7B and Llama8B.\footnote{Phi3.8B supports only a 4K context window, which limits its applicability to document-level simplification. Training Qwen14B on long documents is challenging given our available computational resources.}

Prior work has shown that LLMs outperform earlier simplification systems on both sentence- and document-level simplification~\citep{kew-etal-2023-bless, wu2025indepthevaluationlargelanguage, qiang2025redefiningsimplicity}. We therefore compare our method against LLM-based baselines. 
LLMs are known to be sensitive to prompting. We evaluate using diverse prompting settings: \textbf{zero-shot (zs), few-shot (fs), and iterative feedback (iter)}. The \textbf{zs} and \textbf{fs} prompts adhere to policies and the editing behaviors observed in human references (see Appendix~\ref{app:prompts_simp}). \textbf{Iter} combines automatic evaluation with self-correction: using the same LLM-Judge and evaluation criteria as in our framework, we first generate feedback on a model's output and then prompt the model to revise its simplification accordingly for two rounds (\textit{Iter~1} and \textit{Iter~2}). Prompts for the \textbf{iter} setting are provided in Figures~\ref{fig:prompt_guide_iter_sent} to \ref{fig:correct_doc} (Appendix). 
We use \textbf{zs} prompts for all fine-tuned models, while applying multiple prompting strategies to the vanilla LLMs. 

\paragraph{Common Baseline}
For both sentence- and document-level simplification, we compare our method (\textbf{PO}) against the following baseline: 
\begin{itemize}
    \item \textbf{Base models (Vanilla)}: Instruction-tuned LLMs used directly without additional training. We use \textbf{zs}, \textbf{fs}, and \textbf{iter} prompts.  
    \item \textbf{GPT-5}: A state-of-the-art proprietary LLM that performs strong on simplification~\citep{wu2025indepthevaluationlargelanguage}. We use both \textbf{zs} and \textbf{fs}.
    \item \textbf{SFT on human-written corpora (Parallel)}:  Models fine-tuned on policy-aligned parallel corpora written by human, representing the scenario where such data is available.  We use dev sets of Turk and ASSET (size: $2k$) for Sent-Lexical and Sent-Overall, respectively.\footnote{We used only one reference per sentence, as preliminary experiments showed performance degradation with multiple references during training.} For Doc-Elaboration, we use the remaining $1.7$k parallel data from Newsela. We do not include the same setting for Doc-Lexical because no training data is available.
\end{itemize}

\paragraph{Sentence-level Baseline} 
To investigate the effect of our edit-policy-oriented preference optimization corpus created by LLM-Judge, we compare to LENS-SALSA Preference Optimization \textbf{(LENS\_SALSA)}. LENS-SALSA~\citep{heineman-etal-2023-dancing} is a simplification metric trained on large-scale fine-grained human annotations to approximate LENS scores in a reference-free manner. We use the LENS-SALSA score to construct the preference dataset; candidates with the highest scores are treated as preferred, whereas those with the lowest scores are treated as dispreferred.

\paragraph{Document-level Baseline}

For document-level, we further compare to \textbf{AgentSimp} \citep{fang-etal-2025-collaborative}, an LLM-based multi-agent framework for document simplification built on ChatGPT and reported to outperform the ChatGPT baseline.\footnote{Because the code and data are not publicly available, we replicate PL$^\dagger$, the best-performing setting reported in the paper, using the algorithms and prompts described by the authors.}





\begin{table}[t]
\centering
\small
\resizebox{0.9\columnwidth}{!}{ 
    \begin{tabular}{l l c c}
    \hline
    Model & Method & Turk-SARI & ASSET-LENS \\
    \hline
    \multirow{1}{*}{GPT-5}
    & zs                & $41.2$ & $68.1$ \\
    & fs                & $42.2$ & $70.7$ \\
    \hline
    \multirow{5}{*}{Phi3.8B}
    & Vanilla-zs              & $35.5$ & $68.9$ \\
    & Vanilla-fs              & $\underline{42.2}$ & $70.0$ \\
    & Vanilla-iter1              & $38.9$ & $66.6$ \\
    & Vanilla-iter2              & $39.8$ & $66.0$ \\
    & Parallel             & $36.7$ & $69.6$ \\
    & LENS\_SALSA          & $41.8$ & $\bm{73.5}$ \\
    & PO (Ours)     & $\bm{43.3}$ & $\underline{71.4}$ \\
    \hline
    \multirow{5}{*}{Qwen7B}
    & Vanilla-zs         & $38.0$ & $65.0$ \\
    & Vanilla-fs              & $41.0$ & $68.4$ \\
    & Vanilla-iter1              & $39.7 $ & $63.7$ \\
    & Vanilla-iter2              & $40.6$ & $63.6$ \\
    & Parallel             & $\underline{42.4}$ & $64.9$ \\
    & LENS\_SALSA          & $41.5$ & $\bm{72.5}$ \\
    & PO (Ours)     & $\bm{43.7}$ & $\underline{70.8}$ \\
    \hline
    \multirow{5}{*}{Llama8B}
    & Vanilla-zs               & $39.2$ & $66.9$ \\
    & Vanilla-fs              & $40.7$ & $70.0$ \\
    & Vanilla-iter1              & $42.4 $ & $64.3$ \\
    & Vanilla-iter2              & $\underline{42.8}$ & $64.0$ \\
    & Parallel             & $35.4$ & $68.4$ \\
    & LENS\_SALSA          & $41.7$ & $\bm{73.0}$ \\
    & PO (Ours)     & $\bm{43.7}$ & $\underline{70.2}$ \\
    \hline
    \multirow{5}{*}{Qwen14B}
    & Vanilla-zs             & $38.7$ & $68.0$ \\
    & Vanilla-fs              & $\underline{42.6}$ & $66.5$ \\
    & Vanilla-iter1              & $41.8$ & $64.8$ \\
    & Vanilla-iter2              & $42.3$ & $65.7$ \\
    & Parallel             & $40.4$ & $68.4$ \\
    & LENS\_SALSA          & $42.2$ & $\bm{72.4}$ \\
    & PO (Ours)     & $\bm{43.9}$ & $\underline{70.3}$ \\
    \hline
    \end{tabular}
}
\caption{Sentence-level results (\textbf{bold}: the best under each model group, \underline{underline}: the second best.)}
\label{tab:auto_eval_sent}
\end{table}

\begin{table}[t]
\centering
\small

\begin{subtable}{\columnwidth}
\centering
\resizebox{0.9\columnwidth}{!}{
\begin{tabular}{l l cc cc}
\hline
& & \multicolumn{2}{c}{OSE} & \multicolumn{2}{c}{Newsela} \\
Model & Method & LENS & BERT & LENS & BERT \\
\hline
\multirow{1}{*}{GPT-5}
& zs & $54.7$ & $59.4$ & $33.7$ & $25.4$ \\
& fs & $52.4$ & $56.1$ & $34.9$ & $28.0$ \\
\hline
\multirow{1}{*}{Agent-Simp}
& -- & $34.0$ & $16.4$ & $44.3$ & $19.9$ \\
\hline
\multirow{3}{*}{Qwen7B}
& Vanilla-zs & $41.6$ & $27.3$ & $33.7$ & $20.6$ \\
& Vanilla-fs & $\underline{47.0}$ & $\underline{40.0}$ & $39.6$ & $34.0$ \\
& Vanilla-iter1 & $42.2$ & $28.8$ & $34.7$ & $22.6$ \\
& Vanilla-iter2 & $42.3$ & $29.4$ & $34.8$ & $23.2$ \\
& Parallel & $-$ & $-$ & $\bm{50.9}$ & $\underline{45.4}$ \\
& PO (Ours) & $\bm{53.7}$ & $\bm{50.1}$ & $\underline{50.7}$ & $\bm{52.0}$ \\
\hline
\multirow{3}{*}{Llama8B}
& Vanilla-zs & $47.5$ & $35.0$ & $34.7$ & $25.1$ \\
& Vanilla-fs & $44.3$ & $33.6$ & $39.5$ & $33.2$ \\
& Vanilla-iter1 & $47.5$ & $35.2$ & $35.4$ & $26.5$ \\
& Vanilla-iter2 & $\underline{47.6}$ & $\underline{35.8}$ & $36.2$ & $28.0$ \\
& Parallel & $-$ & $-$ & $\underline{52.4}$ & $\underline{50.6}$ \\
& PO (Ours) & $\bm{53.9}$ & $\bm{51.5}$ & $\bm{52.6}$ & $\bm{55.6}$ \\
\hline
\end{tabular}
}
\caption{Automatic evaluation results. LENS denotes Agg-LENS, and BERT denotes Agg-BERT.} 
\label{tab:auto_doc_eval_metrics_combined_a}
\end{subtable}

\vspace{0.5em}


\begin{subtable}{0.95\columnwidth}
\centering
\resizebox{\columnwidth}{!}{
\begin{tabular}{l l c c c c}
\hline
Model & Method & OSE-L & OSE-D & News-E & News-D \\
\hline
\multirow{1}{*}{\textit{human ref}}
& -- & $26.9$ & $47.4$ & $16.9$ & $48.2$ \\
\hline
\multirow{1}{*}{GPT-5}
& zs & $+10.4$ & $-21.7$ & $-3.6$ & $+7.0$ \\
& fs & $+7.0$ & $-17.8$ & $-3.9$ & $+5.6$ \\
\hline
\multirow{1}{*}{Agent-Simp}
& -- & $-13.9$ & $+5.9$ & $+16.7$ & $-29.1$ \\
\hline
\multirow{3}{*}{Qwen7B}
& Vanilla-zs & $-11.5$ & $+17.4$ & $-9.0$ & $+24.2$ \\
& Vanilla-fs & $\underline{-9.2}$ & $\underline{+9.5}$ & $-6.4$ & $+16.7$ \\
& Vanilla-Iter1 & $-11.4$ & $+16.1$ & $-8.7$ & $+22.7$ \\
& Vanilla-Iter2 & $-11.5$ & $+14.9$ & $-8.5$ & $+22.0$ \\
& Parallel & $-$ & $-$ & $\bm{-1.5}$ & $\bm{-2.0}$ \\
& PO (Ours) & $\bm{-2.2}$ & $\bm{-4.4}$ & $\underline{-3.0}$ & $\underline{+5.1}$ \\
\hline
\multirow{3}{*}{Llama8B}
& Vanilla-zs & $\underline{-10.6}$ & $+8.0$ & $-6.5$ & $+20.1$ \\
& Vanilla-fs & $-15.1$ & $+15.5$ & $-6.4$ & $+21.0$ \\
& Vanilla-Iter1 & $-10.6 $ & $+6.1$ & $-5.5$ & $+17.7$ \\
& Vanilla-Iter2 & $-10.7$ & $\underline{+5.3}$ & $-4.6$ & $+15.7$ \\
& Parallel & $-$ & $-$ & $\bm{-0.7}$ & $\bm{-2.0}$ \\
& PO (Ours) & $\bm{-2.7}$ & $\bm{-3.6}$ & $\underline{-2.5}$ & $\underline{+2.1}$ \\
\hline
\end{tabular}
}
\caption{Edit percentage difference (\%) relative to the \textit{human ref} values: values closer to zero indicate better performance. Values are computed as model minus reference. L: lexical, D: deletion, E: elaboration}
\label{tab:edit_eval_doc_combined_b}
\end{subtable}

\caption{Document-level results (\textbf{bold}: the best under each model group, \underline{underline}: the second best.)}
\label{tab:combined_doc_tables}
\end{table}

\subsection{Evaluation Results}
\label{sec:auto_results}
Results are reported in Tables~\ref{tab:auto_eval_sent}~and~\ref{tab:combined_doc_tables}. 
Our method yields consistent gains across edit policies, model architectures, model sizes, on both sentence- and document-level simplification.
Simplification examples are in Tables~\ref{tab:simp_examplessent} and~\ref{tab:simp_examplesdoc} (Appendix).

\paragraph{Our method enables smaller LLMs to match or surpass GPT-5.}
Across all policies, our method not only surpasses the vanilla models but also achieves results comparable to or even exceeding GPT-5. At the sentence level, the improvements of Turk-SARI range from $+4.5$ to $+7.8$ points under the same zero-shot setting. Even under the Sent-Overall policy, where LLMs are already strong due to their capacity for diverse edits, our method remains robust with improvements ranging from $+2.3$ to $+5.8$ points. 
Remind that LENS\_SALSA is optimized for LENS score, thus expected to achieve high score on ASSET. 
In contrast, it struggled with the Turk-SARI that requires lexical-oriented simplification.
It is remarkable that our method still achieves comparable performance against LENS\_SALSA on ASSET while outperforms it on Turk. A key advantage of our method is its ability to adapt to different policies simply by changing the LLM-Judge prompts.
The same trend holds at the document level. On Newsela, both PO models outperform GPT-5, with Qwen7B improving over vanilla zero-shot prompting by $+17.0$ and $+31.4$ on Agg-LENS and Agg-BERT, and Llama8B by $+17.9$ and $+30.5$. On OSE, GPT-5 remains the strongest model, but our method substantially narrows the gap: Qwen7B gains $+12.1$ and $+22.8$, while Llama8B improves by $+6.4$ and $+16.5$, bringing both close to GPT-5.

\paragraph{Prompt engineering is not robust.}
Prompt engineering alone does not provide consistent gains. We found that sometimes few-shot prompting can degrade performance: for example, $-1.6$ LENS for Qwen14B on ASSET, $-3.2$ Agg-LENS for Llama8B on OSE. The gains from few-shot prompting also vary across models: For example, on Turk, it improves Phi3-3.8B by $6.7$ SARI but Llama8B by only $1.5$. On OSE, it improves Qwen7B by $5.4$ points on Agg-LENS, but degrades Llama8B. These suggest that in-context examples can introduce model-specific biases rather than providing reliable guidance. Iterative feedback is also unstable. At the sentence level, it can improve Turk-SARI, but consistently degrades ASSET-LENS. We find that iterative feedback tends to make models more conservative, leading to fewer meaningful simplification edits. At the document level, iterative feedback brings little to no improvement for either policy. In contrast, after our preference training, all models reach consistently strong performance.

\paragraph{LLM-Judge outperforms crowd-sourced parallel corpus.}
Our method surpasses SFT trained on the crowd-sourced Turk and ASSET parallel corpora at the sentence level, while remaining comparable with SFT trained on the expert-written Newsela corpus. Two factors contribute: (1) \textit{Scalability:} LLM-Judge enables creating large-scale preference data easily and efficiently, whereas SFT is constrained by the scale of human efforts. (2) \textit{Quality Control:} Human references are not always consistent and may even be surpassed by advanced LLMs~\citep{xu2024contrastive, liu-etal-2024-learning, qiang2025redefiningsimplicity}. 
Appendix~\ref{app:train_size} further analyzes the effects of data sizes. Even when trained on the same number of samples ($2$k / $1.5$k), our method consistently outperforms the Parallel baseline. 


\paragraph{PO aligns edit distributions more closely with human references and reduces over-deletion.}
On both OSE and Newsela, our models achieve edit ratios that closely match those of the human references. These results indicate that our method not only improves document-level simplification quality but also effectively steers model behavior toward the desired edit policy. Consistent with prior work~\citep{nandiraju2025automated, fang2026progressive}, vanilla LLMs tend to over-rely on deletion at the document level. Our method substantially alleviates this tendency. Under the same zero-shot prompting setting, on OSE, the deletion rate drops from $64.8\%$ to $43.0\%$ for Qwen7B and from $55.5\%$ to $43.8\%$ for Llama8B, moving both closer to the human reference of $47.4\%$. On Newsela, it decreases from $72.4\%$ to $53.3\%$ for Qwen7B and from $68.3\%$ to $50.3\%$ for Llama8B, again approaching the reference value of $48.2\%$. These improvements are also substantially larger than those achieved by prompting strategies.

\begin{table}[t]
    \centering
    \small




    \begin{subtable}{0.8\columnwidth}
        \centering
        \resizebox{\columnwidth}{!}{
        \begin{tabular}{lccc}
        \hline
        Model & Lex\_P & Lex\_R & Lex\_F1 \\
        \hline
        GPT-5-zs      & $0.21$ $(0.09)$ & $0.36$ $(0.15)$ & $0.27$ $(0.11)$ \\
        GPT-5-fs      & $0.21$ $(0.08)$ & $0.33$ $(0.12)$ & $0.26$ $(0.10)$ \\
        Agent-Simp    & $0.22$ $(0.08)$ & $0.12$ $(0.05)$ & $0.16$ $(0.06)$ \\
        Qwen7B\_PO    & $0.25$ $(0.11)$ & $0.24$ $(0.11)$ & $0.24$ $(0.11)$ \\
        Llama8B\_PO   & $0.24$ $(0.11)$ & $0.21$ $(0.10)$ & $0.22$ $(0.10)$ \\
        \hline
        \end{tabular}
        }
        \caption{OSE}
    \end{subtable}

    \vspace{0.5em}

    \begin{subtable}{0.8\columnwidth}
        \centering
        \resizebox{\columnwidth}{!}{
        \begin{tabular}{lccc}
        \hline
        Model & Ela\_P & Ela\_R & Ela\_F1 \\
        \hline
        GPT-5-zs      & $0.06$ $(0.02)$ & $0.08$ $(0.02)$ & $0.07$ $(0.02)$ \\
        GPT-5-fs      & $0.06$ $(0.01)$ & $0.07$ $(0.02)$ & $0.07$ $(0.02)$ \\
        Agent-Simp    & $0.04$ $(0.01)$ & $0.15$ $(0.03)$ & $0.07$ $(0.02)$ \\
        Qwen7B\_PO    & $0.11$ $(0.03)$ & $0.08$ $(0.02)$ & $0.09$ $(0.02)$ \\
        Llama8B\_PO   & $0.11$ $(0.03)$ & $0.07$ $(0.02)$ & $0.08$ $(0.03)$ \\
        \hline
        \end{tabular}
        }
        \caption{Newsela}
    \end{subtable}
    \caption{Precision, recall, and F1 scores of edit patterns (Lex: lexical edits, Ela:  elaboration edits). Scores in parentheses are exact-matching.}
    \label{tab:ose_newsela_prf}
\end{table}

\subsection{Further Analysis}
\label{sec:edit_pattern}
We evaluate LLMs' edit patterns by computing precision, recall, and F1 against human-edit spans.\footnote{We evaluate in a single-reference setting (see Section~\ref{sec:eval_data}), which may have lowered precision, recall, and F1 scores.}
Because semantically equivalent edits sometimes differ in span boundaries (e.g., `air force' versus `air force fields'), we report a soft-matching, in which two spans are considered as a match when at least $50\%$ of the words overlap in both spans, in addition to exact matching. Appendix~\ref{app:edit_patterns} describes details of edit span recognition.

Table~\ref{tab:ose_newsela_prf} shows that our method achieves competitive F1 scores against GPT-5; however, elaboration is clearly more challenging than lexical editing, as reflected in the lower scores on Newsela. This trend is also consistent with the comparison against the parallel baseline in Table~\ref{tab:combined_doc_tables}; under the Doc-Elaboration policy, our method does not outperform the baseline trained on the professionally edited Newsela corpus.
Elaboration requires inferring what information to add and how much explanation is appropriate. Although Agent-Simp introduces a general-purpose explainer agent, we find that it often over-explains and adds unnecessary details. 
Together, these results suggest that effective elaboration requires stronger awareness of the target reader and more explicit modeling of pedagogical judgment.

In addition to the analysis above, experiments in Appendix~\ref{app:ood} demonstrate that our method transfers effectively across domains. 
Appendix~\ref{app:ablation_study} provides a detailed ablation study.

\section{Conclusion and Future Work}
We propose a framework for adapting text simplification to various edit policies, which is critical for real-world applications. By leveraging LLM-Judge, our method consistently enhances the policy alignment of small open-source LLMs, achieving comparable or even higher performance than the large proprietary LLMs.

Future work could include a thorough human evaluation to assess whether LLM-generated simplifications meet educational expectations in practice.
Moreover, elaboration remains challenging as discussed above. Future work could extend our framework to support finer-grained control by conditioning the LLM judges on target-audience characteristics or reading goals and incorporating expert knowledge into the judging process.

\section*{Limitations}
First, although we designed several simplification policies for L2 education based on prior work in second language acquisition, their educational effectiveness remains to be validated in real learning settings. In particular, it is necessary to deploy these policies in real educational use and examine their long-term impact on learners' comprehension, vocabulary growth, and reading development.

Second, as discussed in the paper, although our method can steer model editing behavior toward the target policy, a gap still remains between model edits and human edits, especially for elaboration. This gap likely stems from the inherent natures of LLMs, which may not always align with how human experts would perform. Future work could therefore explore more fine-grained control over elaboration, for example, by conditioning on learners' knowledge background.


\section*{Acknowledgement}
This work was supported by NTT, Inc. and JSPS KAKENHI Grant Number JP25K03233.

\bibliography{custom}

\appendix

\section{Appendix}
\label{sec:appendix}

\subsection{Preliminaries: Preference Optimization}
\label{app:preliminary_po}
Preference Optimization (PO) was first popularized by InstructGPT~\citep{ouyang2022traininglanguagemodelsfollow} using proximal policy optimization (PPO)~\citep{schulman2017proximalpolicyoptimizationalgorithms}. However, PPO suffers from instability, high variance, and complexity, as it requires a reward model and online reinforcement learning. To address these limitations, Direct Preference Optimization (DPO)~\citep{rafailov2023direct} was proposed as a lightweight alternative, removing the need for an explicit reward model. 

Building on DPO, Contrastive Preference Optimization (CPO)~\citep{xu2024contrastive} further improves memory efficiency through a simpler preference loss combined with a behavior cloning loss, reducing reliance on a reference model required by DPO. CPO has been shown to be effective for machine translation. In parallel, Simple Preference Optimization (SimPO)~\citep{meng2024simpo} also offers a reference-free but more stable formulation, incorporating length normalization and a target reward margin. CPO and SimPO can be combined to CPO-SimPO for improved performance and stability. 

In this study, we use ARPO, which is a variant of CPO designed to mitigate its tendency to overly penalize dispreferred responses that are only marginally worse than preferred ones. Following CPO-SimPO, we also incorporate SimPO into the ARPO loss. We implemented based on the authors' implementation\footnote{\url{https://github.com/fe1ixxu/ALMA/tree/master}} and paper~\citep{xu2025xalma}. Formally, given a policy model $\pi_{\theta}$ and a preference dataset $\mathcal{D}$ consisting of source texts $x$, preferred candidates $y_{w}$, and dispreferred candidates $y_{l}$, we optimize the objective:

{\small
\begin{align}
\mathcal{L}
&= -\mathbb{E}_{(x, y_w, y_l) \sim \mathcal{D}}
\left[
\mathcal{L}_{\text{pref}}(x, y_w, y_l)
+ \mathcal{L}_{\text{BC}}(x, y_w)
\right]
\label{formula:arpo}
\end{align}

\begin{align}
\mathcal{L}_{\text{pref}}(x, y_w, y_l)
&=
\log \sigma \Big(
\frac{\beta}{|y_w|}\log \pi_\theta(y_w \mid x)
\nonumber\\
&\quad
-\tau_\theta(y_w, y_l)\cdot
\frac{\beta}{|y_l|}\log \pi_\theta(y_l \mid x)
-\gamma
\Big)
\label{formula:arpo-pref}
\end{align}

\begin{align}
\mathcal{L}_{\text{BC}}(x, y_w)
&=
\alpha \log \pi_\theta(y_w \mid x)
\label{formula:arpo-bc}
\end{align}
}
where $\beta$ is a hyperparameter, $\sigma$ is the sigmoid function. $\tau_\theta$ is determined by the similarity between $y_w$ and $y_l$, ranging from $0$ to $1$. $|y_w|$ and $|y_l|$ represent the length normalization for the preferred and dispreferred candidates, respectively. 
$\gamma$ denotes the target reward margin, and $\mathcal{L}_{\text{BC}}(x, y_w)$ denotes the behavior cloning loss.

\subsection{Experiment Setting}
\label{app:exp_set}
We used LoRA~\citep{hu2022lora} for both PO and SFT training, with $\alpha=32$ and $r = 16$. For PO, we set the learning rate to $1\mathrm{e}{-4}$. Total batch size was set to $128$ for the sentence level and $4$ for the document level. We set $\beta = 0.1$ and $\gamma = 1.5$ for the SimPO loss~\citep{meng2024simpo}, while $\alpha$ is fixed to $1$. For SFT, we used the LLaMA-Factory package~\citep{zheng2024llamafactory}. We set the learning rate to $2e-4$ for sentence-level training with $1$ epoch, and to $5e-5$ for document-level training with $3$ epochs. 

We used the vLLM package~\citep{kwon2023vllm}\footnote{\url{https://github.com/vllm-project/vllm}} for inference with open-source LLMs. We set the decoding parameters to temperature $=0$, top-$p=1.0$, and top-$k=-1$. For Qwen3-32B in thinking mode, which was used only for LLM-Judge, we followed the official settings\footnote{\url{https://huggingface.co/Qwen/Qwen3-32B}}
 with temperature $=0.6$ and top-$p=0.95$. For the document-level SFT models, we also used temperature $=0.6$ and top-$p=0.95$ to avoid repetitive outputs. Sentence-level experiments are on a single NVIDIA A6000 Ada 48GB GPU. Document-level experiments are on a single NVIDIA H100 SXM5 94GB GPU.

For OTAlign, we used the supervised setting of the authors' implementation\footnote{\url{https://github.com/yukiar/OTAlign}}, using $\tau = 0.88$ and a threshold of $0.40$. SARI was computed with the EASSE package~\citep{alva-manchego-etal-2019-easse}. LENS, LENS-SALSA\footnote{\url{https://github.com/Yao-Dou/LENS}}, Agg-LENS, and Agg-BERT were computed with the authors’ implementation.\footnote{\url{https://github.com/cardiffnlp/document-simplification}}

\subsection{Implementation of Edit Analysis}

In this section, we describe our implementation of the edit-category distribution and edit-pattern analysis. All implementations are based on the edit classifier from~\citet{laban-etal-2023-swipe}.\footnote{\url{https://github.com/salesforce/simplification}}

\subsubsection{Edit-category Distribution}
\label{app:edit_dist}
\begin{figure*}[t]
  \centering
  \includegraphics[width=0.9\linewidth]{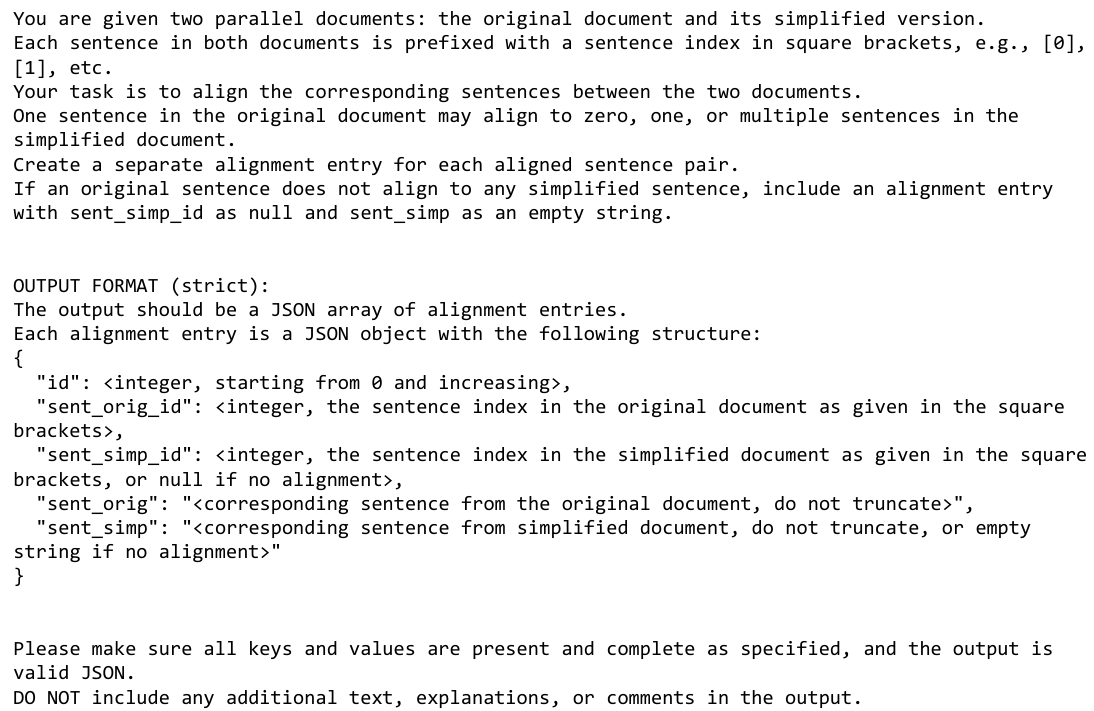}
    \caption{System messages for sentence alignment}
  \label{fig:align_sys}
\end{figure*}

\begin{figure*}[t]
  \centering
  \begin{subfigure}{\linewidth}
    \centering
    \includegraphics[width=0.7\linewidth]{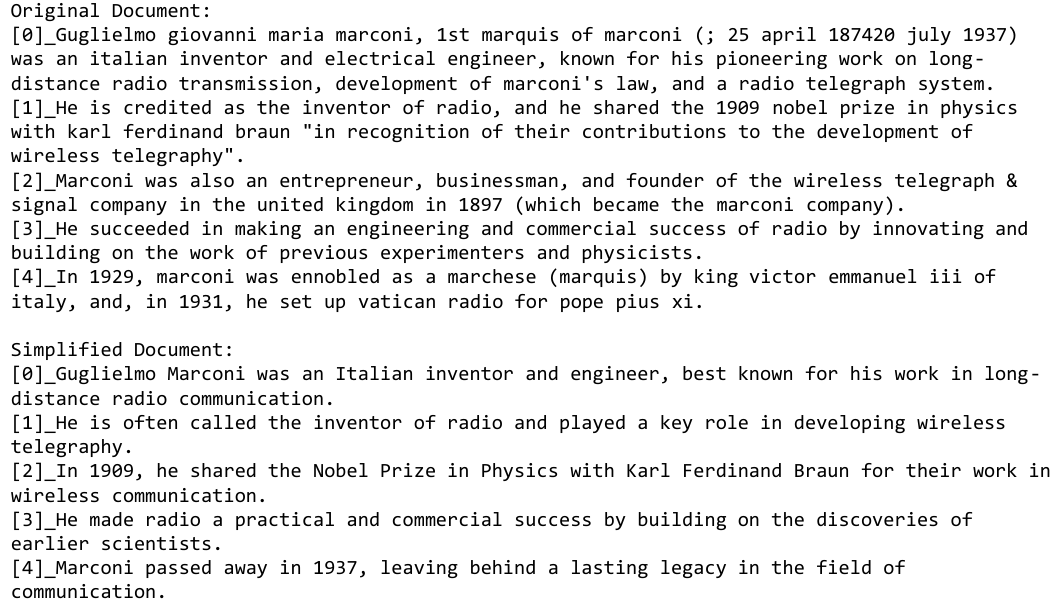}
    \caption{Input}
  \end{subfigure}
  \begin{subfigure}{\linewidth}
    \centering
    \includegraphics[width=0.7\linewidth]{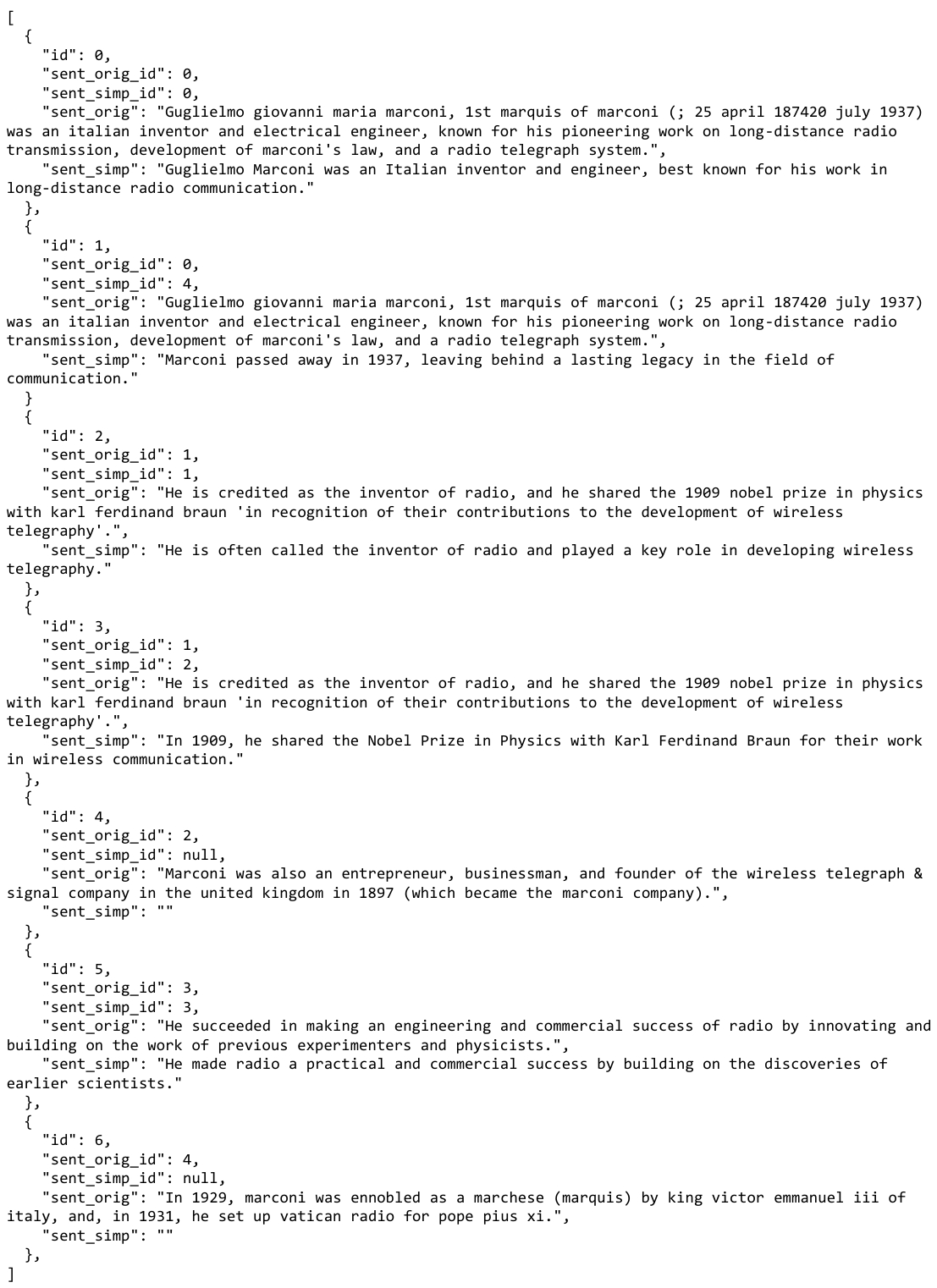}
    \caption{Corresponding output}
  \end{subfigure}
  \caption{One-shot example for sentence alignment}
  \label{fig:align_exp}
\end{figure*}

Given a source document and its simplified version, the edit classifier assigns each edit to one of $19$ fine-grained categories (see the original paper for details). We group the policy-relevant categories into three broader edit types for analysis: \textit{deletion} (semantic deletion, syntactic deletion, and nonsim noise deletion), \textit{elaboration} (semantic elaboration example, semantic elaboration background, and semantic elaboration generic), and \textit{lexical} (lexical generic and lexical entity). All remaining categories are kept unchanged. The classifier is based on RoBERTa-large~\citep{liu2019roberta} and is limited to inputs of at most $512$ tokens. To process longer documents, we first align source and simplified sentences using gpt-oss-120b~\cite{openai2025gptoss120bgptoss20bmodel} in a one-shot setting (Figures~\ref{fig:align_sys} and~\ref{fig:align_exp}). We then merge aligned sentence pairs into larger chunks whenever they share a source or simplified sentence, and apply the classifier to each chunk. Source sentences with no aligned simplified counterpart are labeled as \textit{deletion}.

Because the classifier predicts span-level edits, raw edit counts are not directly comparable: for instance, deleting a single word and deleting an entire sentence would each count as one edit. To make the statistics comparable, we convert predictions to the word level by assigning each word in a labeled span or deleted sentence its predicted edit category. We then compute the proportion of each category as the number of words assigned to that category divided by the total number of edited words.

\subsubsection{Edit-pattern Analysis}
\label{app:edit_patterns}
Given a source text and its simplification, the authors’ implementation first derives edit operations \textit{insert}, \textit{delete}, and \textit{equal} between the two texts, which indicate how the simplification is performed. These operations are then converted into a marked text sequence, where inserted and deleted spans are wrapped with special boundary tokens. The classifier outputs a category to each inserted or deleted span. 

For comparison, we extract spans as follows: for Elaboration, we use the inserted span in the simplified text; for Deletion, the deleted span in the source text; and for Lexical edits, the deleted span in the source text, since the same complex expression may be simplified in multiple valid ways. For each category, we compare spans from the model output with the human reference and identify correspondences.

\subsection{Prompts}
\label{app:prompts}

\subsubsection{LLM-Judge}
\label{app:prompts_judge}
\begin{figure*}[t]
  \centering
  \includegraphics[width=0.8\linewidth]{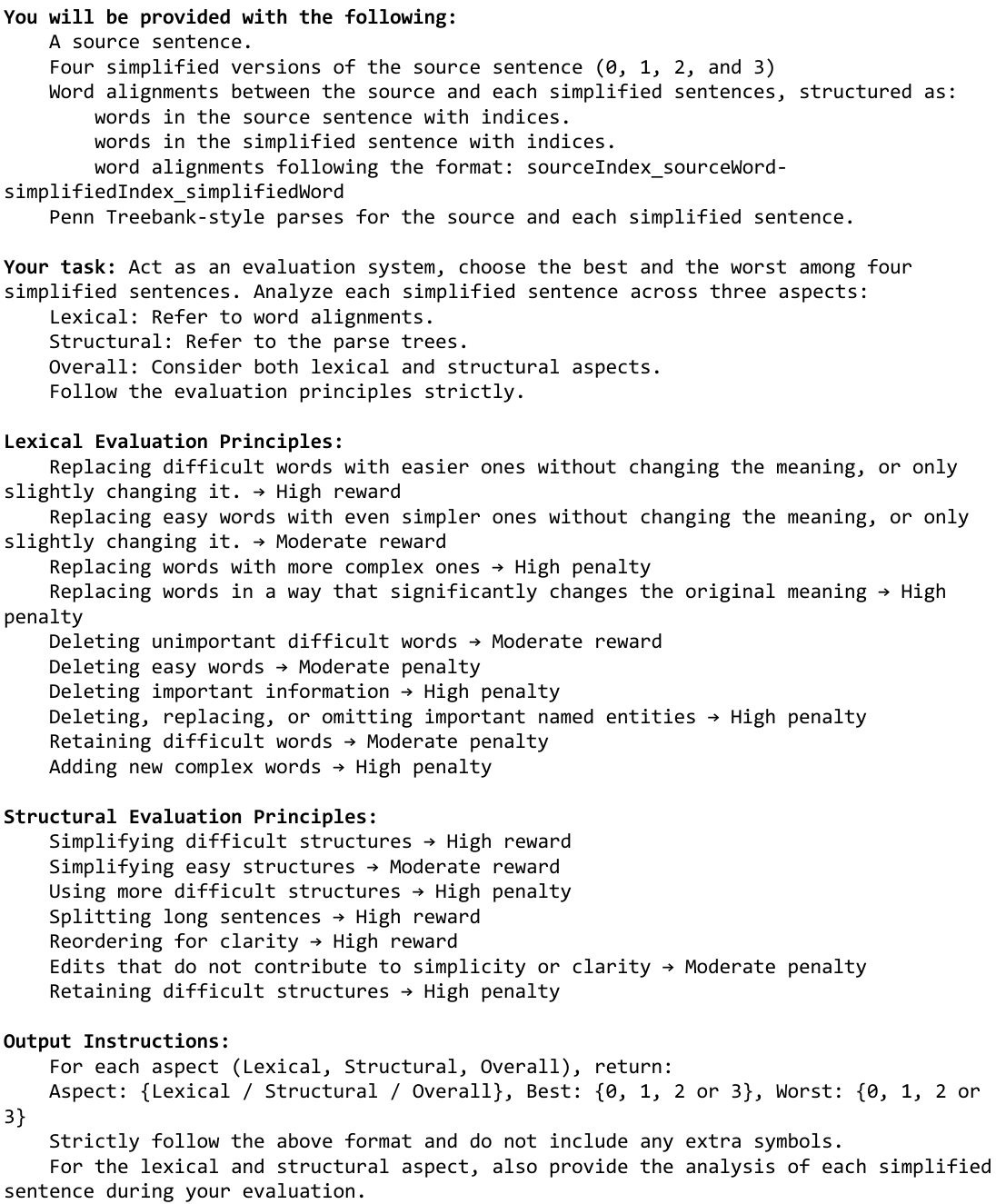}
  \caption{Sentence-level evaluation guidelines provided to the judge LLM}
  \label{fig:guidlines_sent}
\end{figure*}

\begin{figure*}[t]
  \centering
  \begin{subfigure}{\linewidth}
    \centering
    \includegraphics[width=0.7\linewidth]{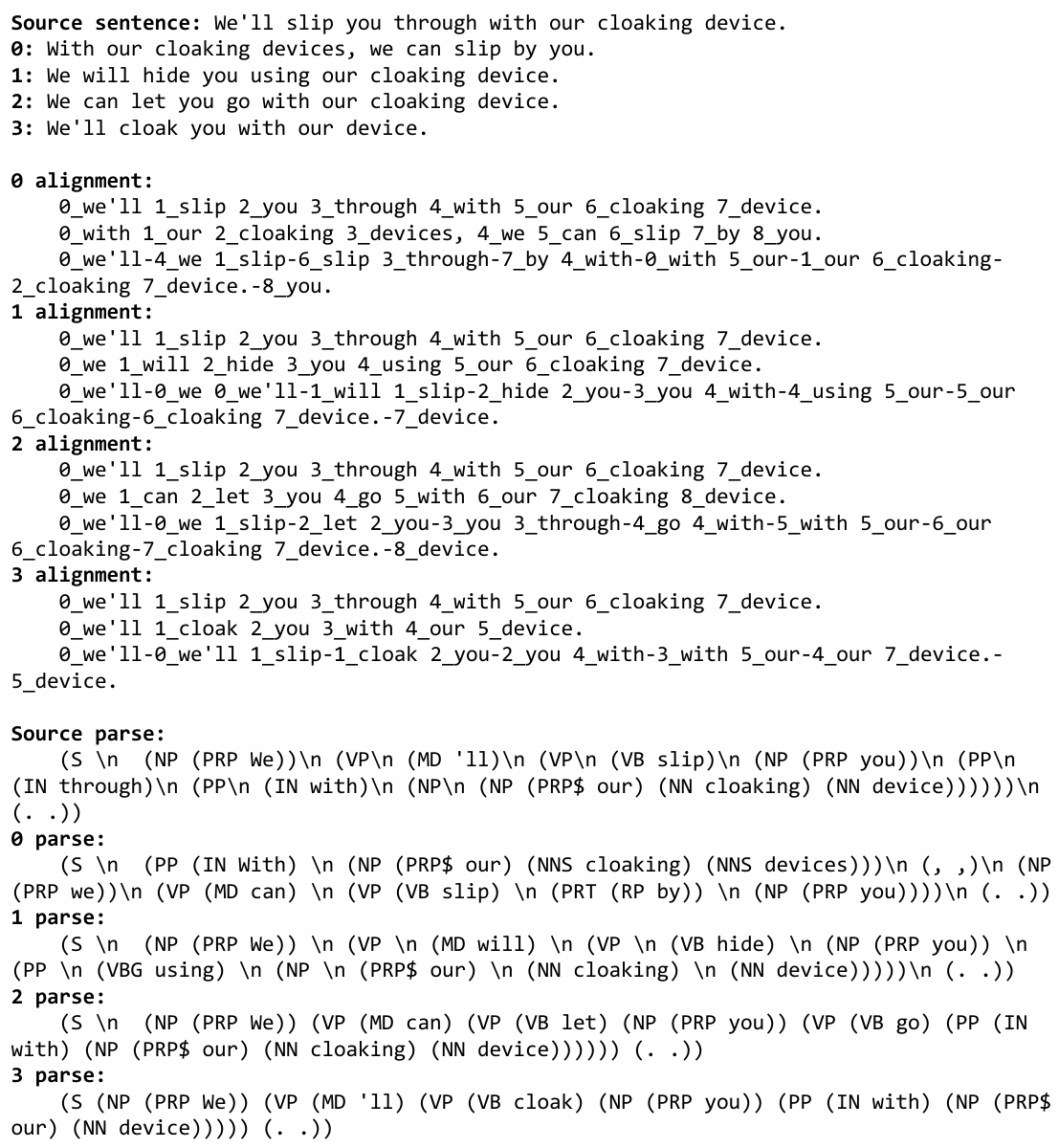}
    \caption{Input}
  \end{subfigure}

  \begin{subfigure}{\linewidth}
    \centering
    \includegraphics[width=0.7\linewidth]{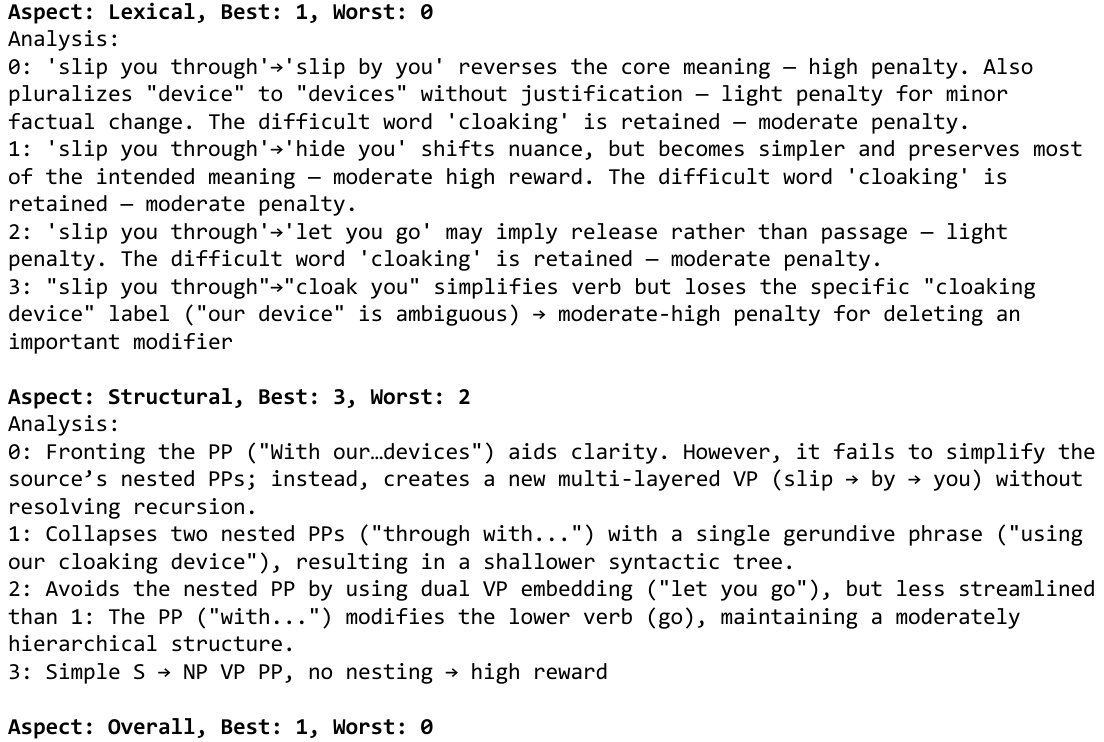}
    \caption{Corresponding output showing evaluations on lexical, structural, and overall dimensions.}
  \end{subfigure}
  \caption{Example 1 for sentence-level LLM-Judge}
  \label{fig:judge_example1_sent}
\end{figure*}

\begin{figure*}[t]
  \centering
  \begin{subfigure}{\linewidth}
    \centering
    \includegraphics[width=0.8\linewidth]{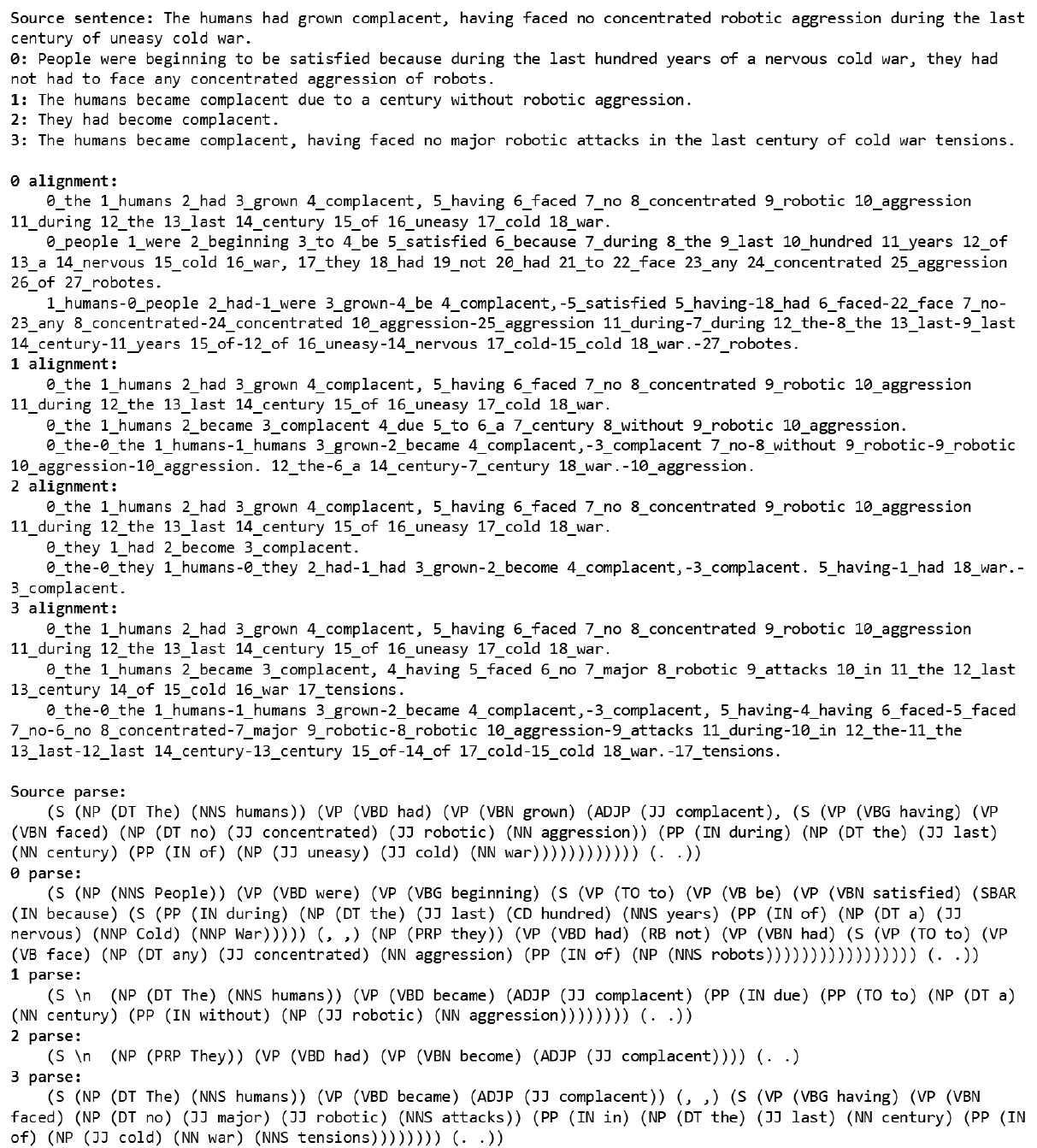}
    \caption{Input}
  \end{subfigure}

  \begin{subfigure}{\linewidth}
    \centering
    \includegraphics[width=0.7\linewidth]{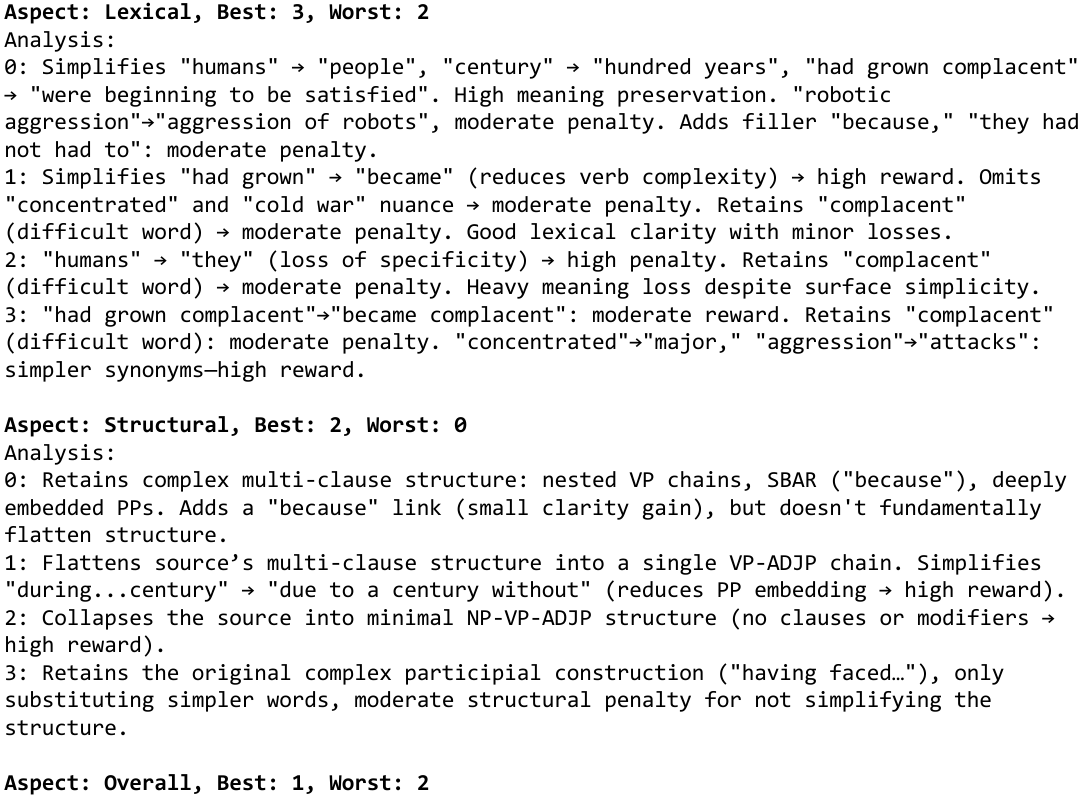}
    \caption{Corresponding output showing evaluations on lexical, structural, and overall dimensions.}
  \end{subfigure}
  \caption{Example 2 for sentence-level LLM-Judge}
  \label{fig:judge_example2_sent}
\end{figure*}

\begin{figure*}[t]
  \centering
  \begin{subfigure}{\linewidth}
    \centering
    \includegraphics[width=0.6\linewidth]{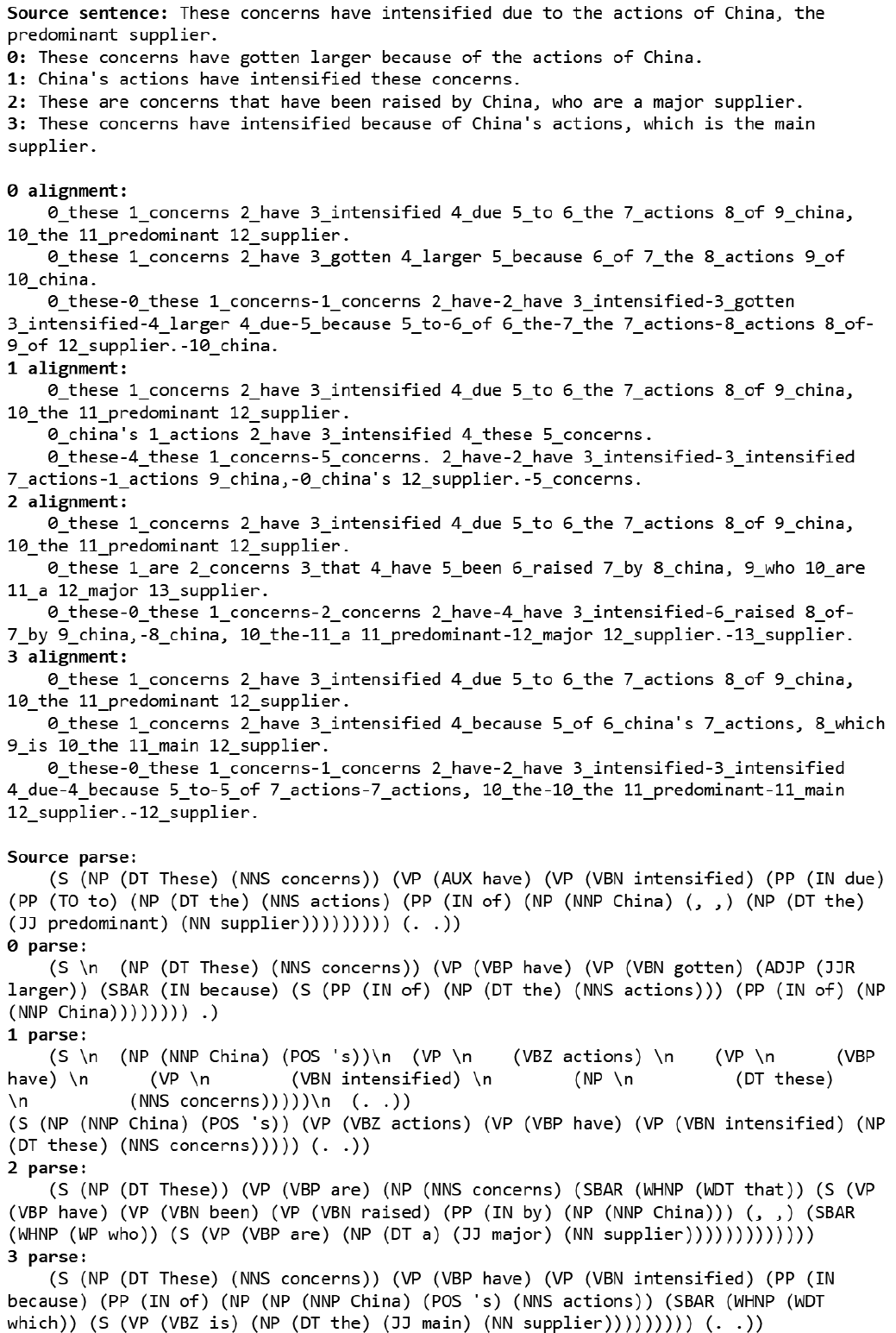}
    \caption{Input}
  \end{subfigure}

  \begin{subfigure}{\linewidth}
    \centering
    \includegraphics[width=0.68\linewidth]{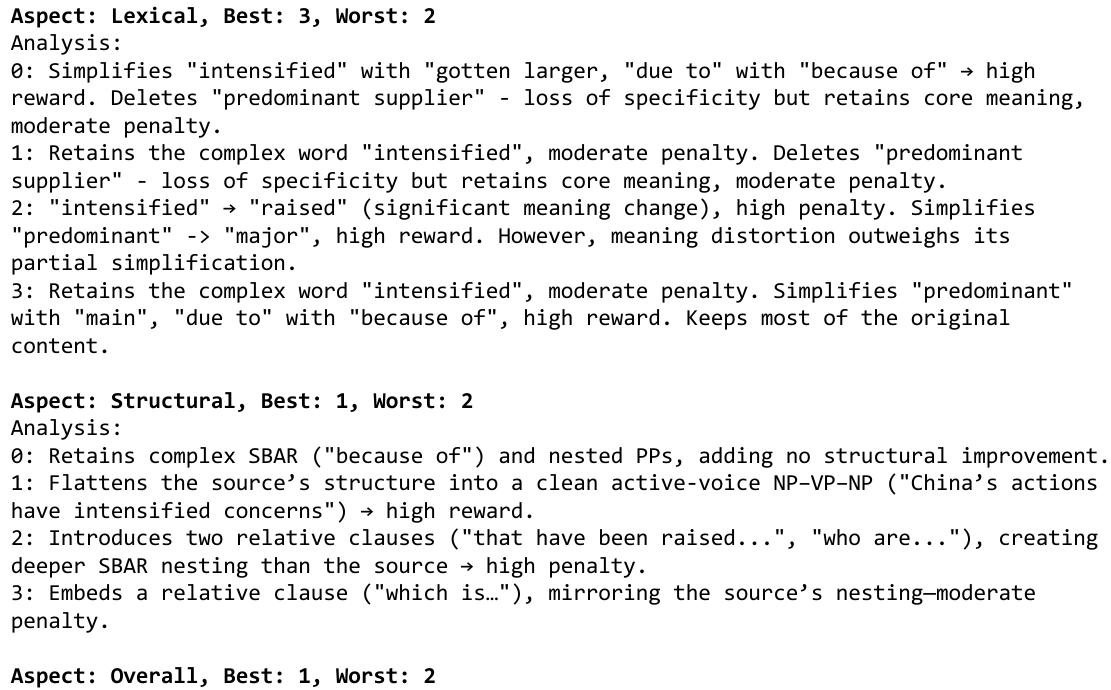}
    \caption{Corresponding output showing evaluations on lexical, structural, and overall dimensions.}
  \end{subfigure}
  \caption{Example 3 for sentence-level LLM-Judge}
  \label{fig:judge_example3_sent}
\end{figure*}

\begin{figure*}[t]
  \centering
  \includegraphics[width=0.8\linewidth]{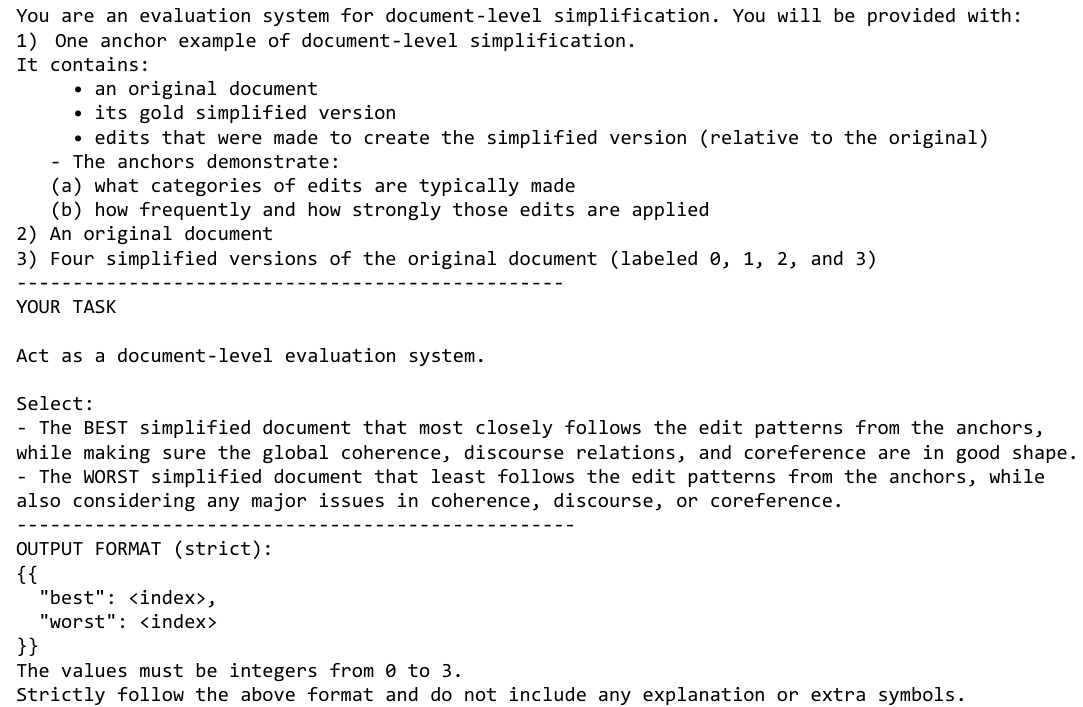}
  \caption{Document-level evaluation guidelines provided to the judge LLM}
  \label{fig:guidlines_doc}
\end{figure*}

Evaluation guidelines for LLM-Judge are provided in the system message of the prompts. They include the task description, the materials provided to the judge, the evaluation principles, and the required output format. The sentence-level and document-level guidelines are shown in Figures~\ref{fig:guidlines_sent} and~\ref{fig:guidlines_doc}, respectively. The sentence-level prompt also includes three examples (Figures~\ref{fig:judge_example1_sent}, \ref{fig:judge_example2_sent}, and~\ref{fig:judge_example3_sent}). In each example, the input contains a source sentence along with four simplification candidates, accompanied by their corresponding word alignments and syntactic parses. The output includes evaluation analysis and decisions across lexical, structural, and overall dimensions, documented by the authors.
For document-level evaluation, the prompt includes one anchor example for each policy (Table~\ref{tab:anchor_ose}). 
Each anchor example contains a source document, a gold simplified document representing the target policy, and edit information derived from sentence alignment and the edit classifier (see~\ref{app:edit_dist}). The judge model was instructed to make its judgments referring to the anchor examples while considering the overall quality of document simplification (e.g., coherence and coreference).

\subsubsection{Simplification Generation}
\label{app:prompts_simp}
\begin{figure*}[t]
  \centering
  \begin{subfigure}{\linewidth}
    \centering
    \includegraphics[width=0.7\linewidth]{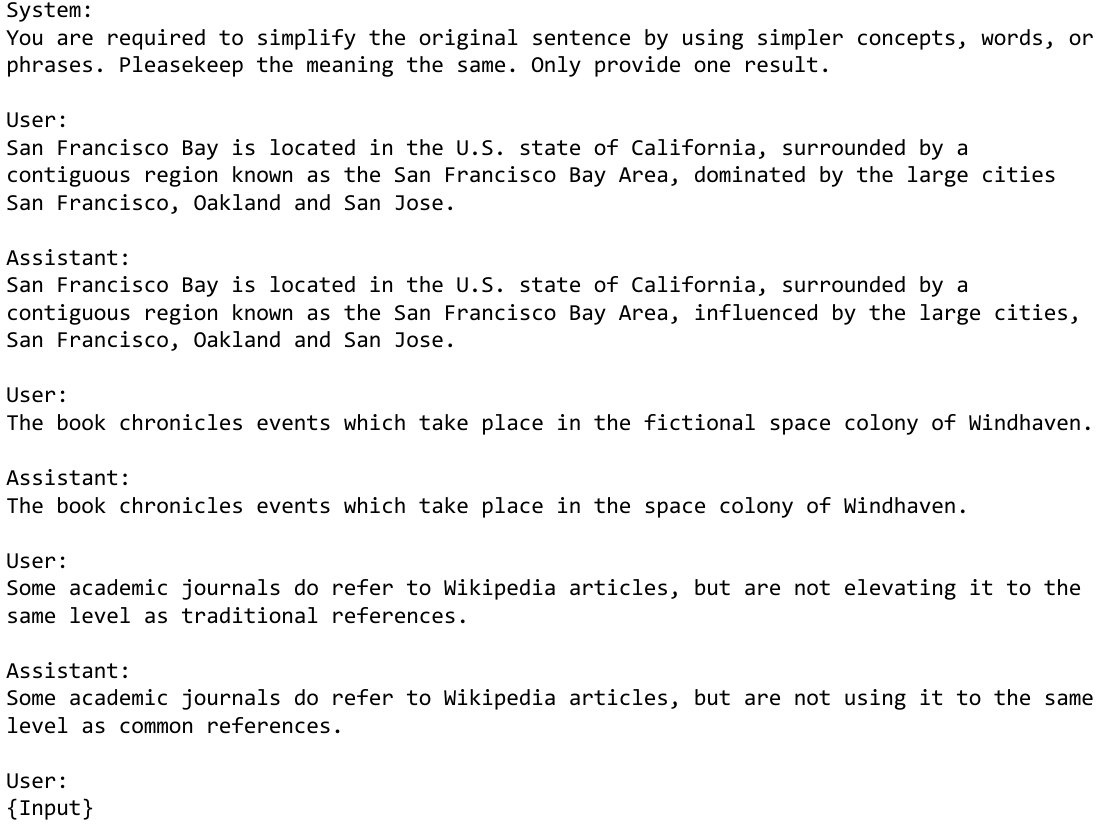}
    \caption{Sent-Lexical}
  \end{subfigure}

  \begin{subfigure}{\linewidth}
    \centering
    \includegraphics[width=0.7\linewidth]{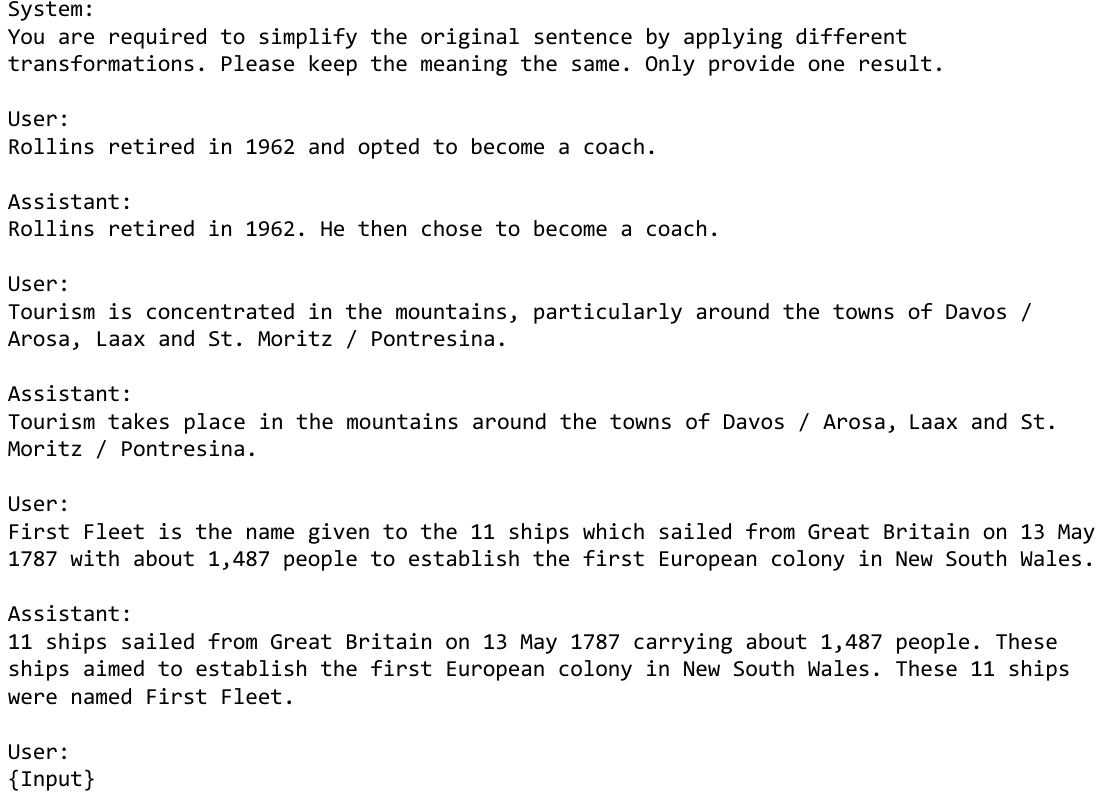}
    \caption{Sent-Overall}
  \end{subfigure}
  \caption{Prompts used for sentence-level simplification generation (from~\citet{wu2025indepthevaluationlargelanguage}). }
  \label{fig:prompt_simp_sent}
\end{figure*}

\begin{figure*}[t]
  \centering
  \begin{subfigure}{\linewidth}
    \centering
    \includegraphics[width=0.8\linewidth]{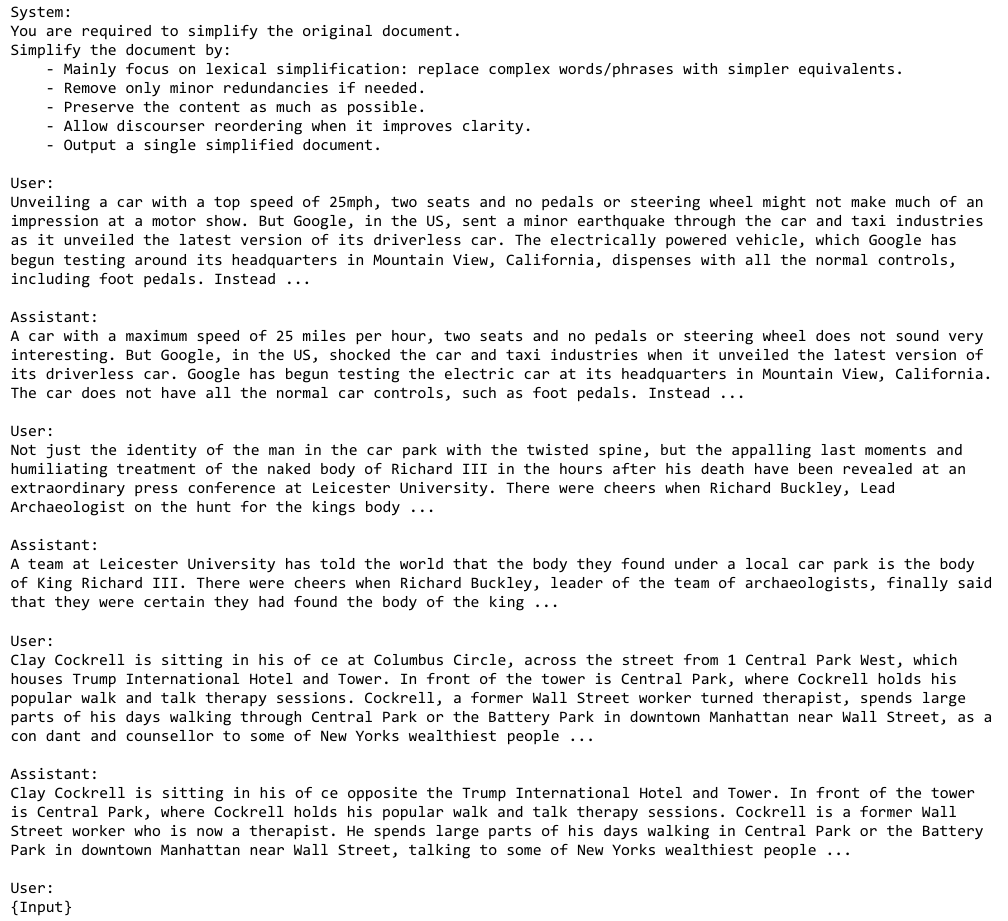}
    \caption{Doc-Lexical}
  \end{subfigure}

  \begin{subfigure}{\linewidth}
    \centering
    \includegraphics[width=0.7\linewidth]{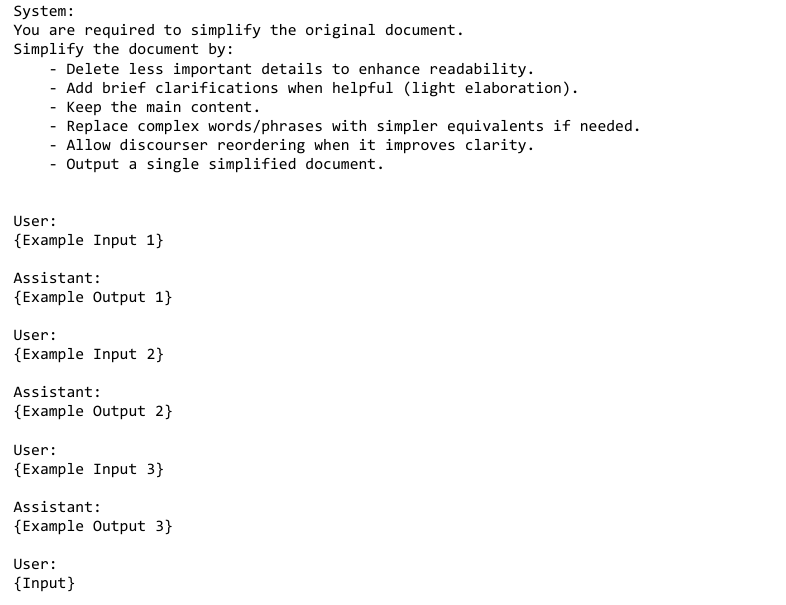}
    \caption{Doc-Elaboration}
  \end{subfigure}
  \caption{Prompts used for document-level simplification generation. The instructions are designed to reflect the reference editing behavior observed in the OSE and Newsela corpora.}
  \label{fig:prompt_simp_doc}
\end{figure*}

We use prompts for simplification generation at three stages: candidate collection, fine-tuning, and evaluation. The sentence-level prompts are from a previous study~\citep{wu2025indepthevaluationlargelanguage} and illustrated in Figure~\ref{fig:prompt_simp_sent}. They are carefully designed and validated on the ASSET and Turk development sets. Each prompt contains an instruction followed by three in-context examples, which we employ for candidate collection. For document-level simplification, we design a new set of prompts, shown in Figure~\ref{fig:prompt_simp_doc}, based on edit behaviours observed in human references from the OSE and Newsela corpora. Specifically, we use the edit classifier to estimate the distribution of edit categories in each dataset. In OSE, deletion is the most frequent edit type, accounting for about 45\% of edits, followed by lexical edits at roughly 25\%. In Newsela, deletion is also the most common edit type, making up around 50\% of edits, followed by elaboration accounts for about 15\%. Based on these patterns, we design prompts that encourage the model to perform the desired edits while still allowing deletion of less important information. For fine-tuning and evaluation on target LLMs, we only use the system instruction part of the prompts.

\subsubsection{Parse}
\label{app:parse}
\begin{figure*}[t]
  \centering
  \includegraphics[width=0.7\linewidth]{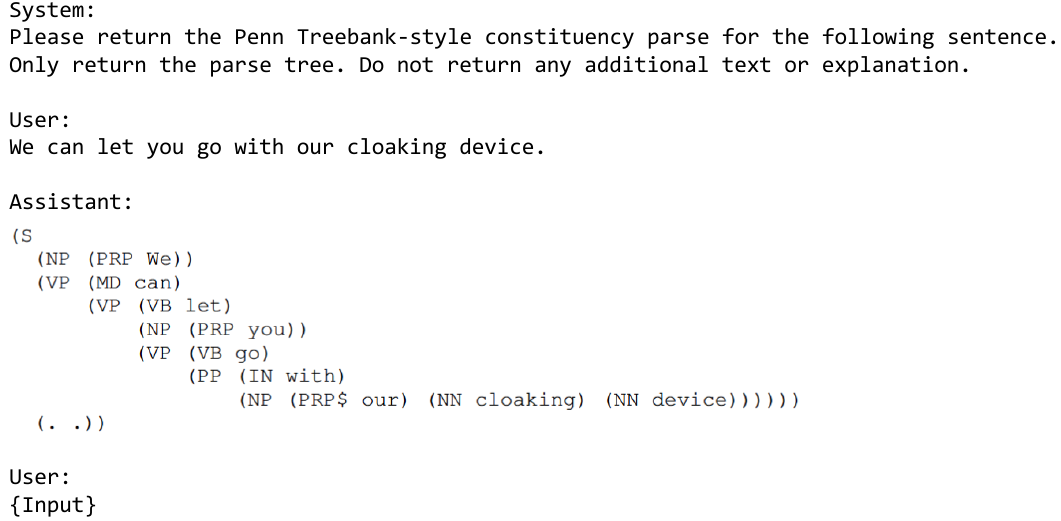}
  \caption{Prompts used for parsing}
  \label{fig:prompt_parse_sent}
\end{figure*}

We use a 1-shot prompt to obtain parses, as shown in Figure~\ref{fig:prompt_parse_sent}. The model receives a system prompt specifying the task and a single input–output demonstration to illustrate the expected format.

\subsection{Ablation Study}
\label{app:ablation_study}

\subsubsection{Evidence-Guided Judge Design}
\label{app:abl_evidence_judge}

As described in Section~\ref{sec:step2judge}, our sentence-level preference construction adopts an evidence-guided Judge design. This design provides the LLM-Judge with human-designed evaluation criteria and external linguistic evidence, including word alignments and parse trees. Rather than asking the judge to output only best/worst labels, we require it to relate observed lexical and syntactic edits to explicit reward or penalty principles. This makes preference-data construction more transparent and traceable, as users can inspect the judge's rationale before the final decision and verify whether the selected preference pairs are consistent with the intended edit policy. 

To examine whether this design contributes to the downstream performance, we conduct an ablation study by removing the detailed reward/penalty criteria, word alignments, and parse information from the judge input. As shown in Table~\ref{tab:guideline_ablation}, models trained on preference data constructed without the evidence-guided design show performance drops, especially on ASSET-LENS. This suggests that evidence-guided judging provides more effective learning signals for preference optimization, particularly for improving overall simplification quality.

Manual inspection of a small number of samples suggests that evidence-guided design helps the judge place more value on edit-level simplicity changes rather than surface-level fidelity to the source. Table~\ref{tab:guideline_analysis_comparison} provides an illustrative example. \textbf{With the evidence-guided design}, the judge explicitly rewards candidate 0 for simplifying the complex construction ``lies in an area known as... characterized by'' into ``is in... where... interact,'' and for replacing ``producing'' with the simpler synonym ``causing.'' It also penalizes candidate 1 for reversing the original subject--location relation and using the less precise verb ``meet.'' As a result, candidate 0 is selected as the best overall candidate and candidate 1 as the worst. \textbf{Without the evidence-guided design}, the judge still detects the major error in candidate 1, but its analysis is less aligned with the intended edit policy. In particular, it favors candidate 2 over candidate 0 mainly because ``frequently'' is closer to the source word ``frequent,'' even though this choice does not make the sentence simpler. This leads to a different choice for the best candidate.

\begin{table}[t]
\centering
\small
\begin{tabular}{llcc}
\toprule
Model & Method & Turk-SARI & ASSET-LENS \\
\midrule
Phi3.8B 
& PO                  & $\bm{43.3}$ & $\bm{71.4}$ \\
& w/o evidence       & $43.3$       & $69.3$ \\
\midrule
Qwen7B  
& PO                  & $\bm{43.7}$ & $\bm{70.8}$ \\
& w/o evidence       & $\bm{43.7}$       & $68.2$ \\
\midrule
Llama8B 
& PO                  & $\bm{43.7}$ & $\bm{70.2}$ \\
& w/o evidence       & $43.6$       & $68.3$ \\
\midrule
Qwen14B 
& PO                  & $\bm{43.9}$ & $\bm{70.3}$ \\
& w/o evidence       & $43.6$       & $67.9$ \\
\bottomrule
\end{tabular}
\caption{Ablation study on the effect of evidence-guided judge design on simplification performance.}
\label{tab:guideline_ablation}
\end{table}

\begin{table*}[t]
\small
\centering
\resizebox{\textwidth}{!}{
\begin{tabular}{p{0.03\textwidth} p{0.3\textwidth} p{0.07\textwidth} p{0.45\textwidth} p{0.35\textwidth}}
\toprule
\multicolumn{5}{p{1.3\textwidth}}{
\textbf{Source:} Most of the state lies in an area known as Tornado Alley characterized by frequent interaction between cold and warm air masses producing severe weather.
} \\
\midrule
\textbf{ID} & \textbf{Candidate} & \textbf{Aspect}
& \textbf{w/ evidence-guided judge}
& \textbf{w/o evidence-guided judge} \\
\midrule

0 &
Most of the state is in Tornado Alley, where cold and warm air masses often interact, causing severe weather. &
Lexical &
Replaces ``lies in an area known as... characterized by'' with ``is in... where... interact'' (active voice simplification) $\to$ high reward. ``Producing'' $\to$ ``causing'' (direct synonym) $\to$ high reward. Retains key terms (``Tornado Alley,'' ``air masses,'' ``severe weather'') $\to$ high reward. &
Uses ``is in Tornado Alley, where cold and warm air masses often interact, causing severe weather.'' The word ``often'' is a good substitute for ``frequent.'' ``Causing'' is correct. Lexically accurate. \\
\cmidrule(lr){3-5}

& &
Structural &
Flattens the source's complex participial clause (``characterized by... producing'') into a relative clause (``where... interact, causing'') $\to$ high reward. &
Structure is similar, uses ``where'' clause. Good. \\
\midrule

1 &
Tornado Alley is located in the state where cold and warm air masses frequently meet and cause severe weather. &
Lexical &
Reverses subject-relationship (``Tornado Alley is located in the state'' vs. original focus on ``most of the state lies in Tornado Alley'') $\to$ moderate penalty. Replaces ``interaction'' with ``meet'' (less precise) $\to$ moderate penalty. &
Changes the structure a bit. Says Tornado Alley is located in the state, which might be a slight misrepresentation. Also, ``meet'' is less precise than ``interact.'' Maybe not as good lexically. \\
\cmidrule(lr){3-5}

& &
Structural &
Inverts subject-object relationship, creating a less intuitive focus (Tornado Alley as subject) $\to$ high penalty. Adds ``located in'' (redundant phrasing) $\to$ moderate penalty. &
Changes the subject to Tornado Alley, which might not follow the original structure as closely. Structural change. \\
\midrule

2 &
Most of the state is in Tornado Alley, where cold and warm air masses frequently interact, causing severe weather. &
Lexical &
Identical to 0 except ``often'' $\to$ ``frequently'' (synonym with no simplification) $\to$ neutral. &
Similar to 0 but uses ``frequently interact'' instead of ``often interact.'' ``Frequently'' is closer to the original's ``frequent.'' So maybe better than 0. \\
\cmidrule(lr){3-5}

& &
Structural &
Same structure as 0 $\to$ high reward. &
Same as 0, maybe better with ``frequently interact.'' \\
\midrule

3 &
Most of the state is in Tornado Alley, an area known for frequent clashes between cold and warm air masses that create severe weather. &
Lexical &
Replaces ``interaction'' with ``clashes'' (slightly more vivid but not simpler) $\to$ moderate penalty. ``Create'' is simpler than ``producing'' $\to$ high reward. &
Uses ``clashes between cold and warm air masses that create severe weather.'' ``Clashes'' is a bit more vivid but might not be as precise as ``interaction.'' ``Create'' is correct, but ``clashes'' could be a lexical issue. \\
\cmidrule(lr){3-5}

& &
Structural &
Adds an appositive phrase (``an area known for...'') that increases nesting slightly $\to$ moderate penalty. &
More complex structure with ``an area known for...'' which adds a bit more. Maybe more complex than needed. \\
\bottomrule
\end{tabular}
}
\caption{
Comparison of lexical and structural judge analyses with and without evidence-guided judge design. 
The analyses are taken from the model output before the final judgments. 
Final judgments (best/worst) for w/ evidence-guided judge were Lexical: 0/1, Structural: 0/1, Overall: 0/1, while for w/o evidence-guided judge were Lexical: 2/1, Structural: 0/1, Overall: 2/1.
}
\label{tab:guideline_analysis_comparison}
\end{table*}

\subsubsection{PO vs. SFT on the Preferred Candidates}
Our PO models are trained with both preferred and dispreferred candidates through preference optimization. To examine whether the preferred responses alone are sufficient for edit-policy alignment, we conduct an additional ablation, SFT\_Chosen, where models are fine-tuned only on the preferred responses from the generated preference pairs. As shown in Table~\ref{tab:po_vs_sft}, SFT\_Chosen achieves strong performance, confirming the quality of the selected preferred candidates. At the sentence level, SFT\_Chosen and PO obtain comparable Turk-SARI scores, while PO shows a modest advantage on ASSET-LENS across all models. At the document level, PO outperforms SFT\_Chosen under both edit policies. Overall, these results suggest that preference optimization can provide more stable gains than training only on preferred candidates, especially for document-level simplification.

\begin{table}[t]
\centering
\small
\begin{subtable}{0.95\columnwidth}
\centering
\begin{tabular}{llcc}
\hline
Model & Method & Turk-SARI & ASSET-LENS \\
\hline
Phi3.8B & SFT\_Chosen & $43.2$ & $69.8$ \\
        & PO  & $\bm{43.3}$ & $\bm{71.4}$ \\
\hline
Qwen7B  & SFT\_Chosen & $\bm{43.7}$ & $70.1$ \\
        & PO  & $\bm{43.7}$ & $\bm{70.8}$ \\
\hline
Llama8B & SFT\_Chosen & $\bm{43.9}$ & $70.0$ \\
        & PO  & $43.7$ & $\bm{70.2}$ \\
\hline
Qwen14B & SFT\_Chosen & $\bm{43.9}$ & $70.2$ \\
        & PO  & $\bm{43.9}$ & $\bm{70.3}$ \\
\hline
\end{tabular}
\caption{Sentence-level}
\end{subtable}

\vspace{0.8em}

\begin{subtable}{\columnwidth}
\centering
\resizebox{0.98\columnwidth}{!}{%
\begin{tabular}{llcccc}
\hline
& & \multicolumn{2}{c}{OSE} & \multicolumn{2}{c}{Newsela} \\
\cline{3-6}
Model & Method & LENS & BERT & LENS & BERT \\
\hline
Qwen7B & SFT\_Chosen & $50.3$ & $42.5$ & $50.5$ & $51.5$ \\
       & PO  & $\bm{53.7}$ & $\bm{50.1}$ & $\bm{50.7}$ & $\bm{52.0}$ \\
\hline
Llama8B & SFT\_Chosen & $\bm{54.6}$ & $49.6$ & $51.7$ & $51.9$ \\
        & PO  & $53.9$ & $\bm{51.5}$ & $\bm{52.6}$ & $\bm{55.6}$ \\
\hline
\end{tabular}%
}
\caption{Document-level}
\end{subtable}

\caption{Comparison of preference optimization and SFT on our preference data.}
\label{tab:po_vs_sft}
\end{table}

\subsubsection{Training Corpus Size vs. Performance}
\label{app:train_size}

We investigate how the size of the preference optimization dataset affects model performance. For sentence-level simplification, we sample subsets of $1$k, $2$k, $3$k, and $5$k preference pairs from the full $7$k training set. For document-level simplification, we sample subsets of $1.5$k and $2.5$k preference pairs from the full $3.5$k training set. All other training settings are kept fixed.

\begin{figure*}[t]
  \small
  \centering
  \includegraphics[width=0.85\linewidth]{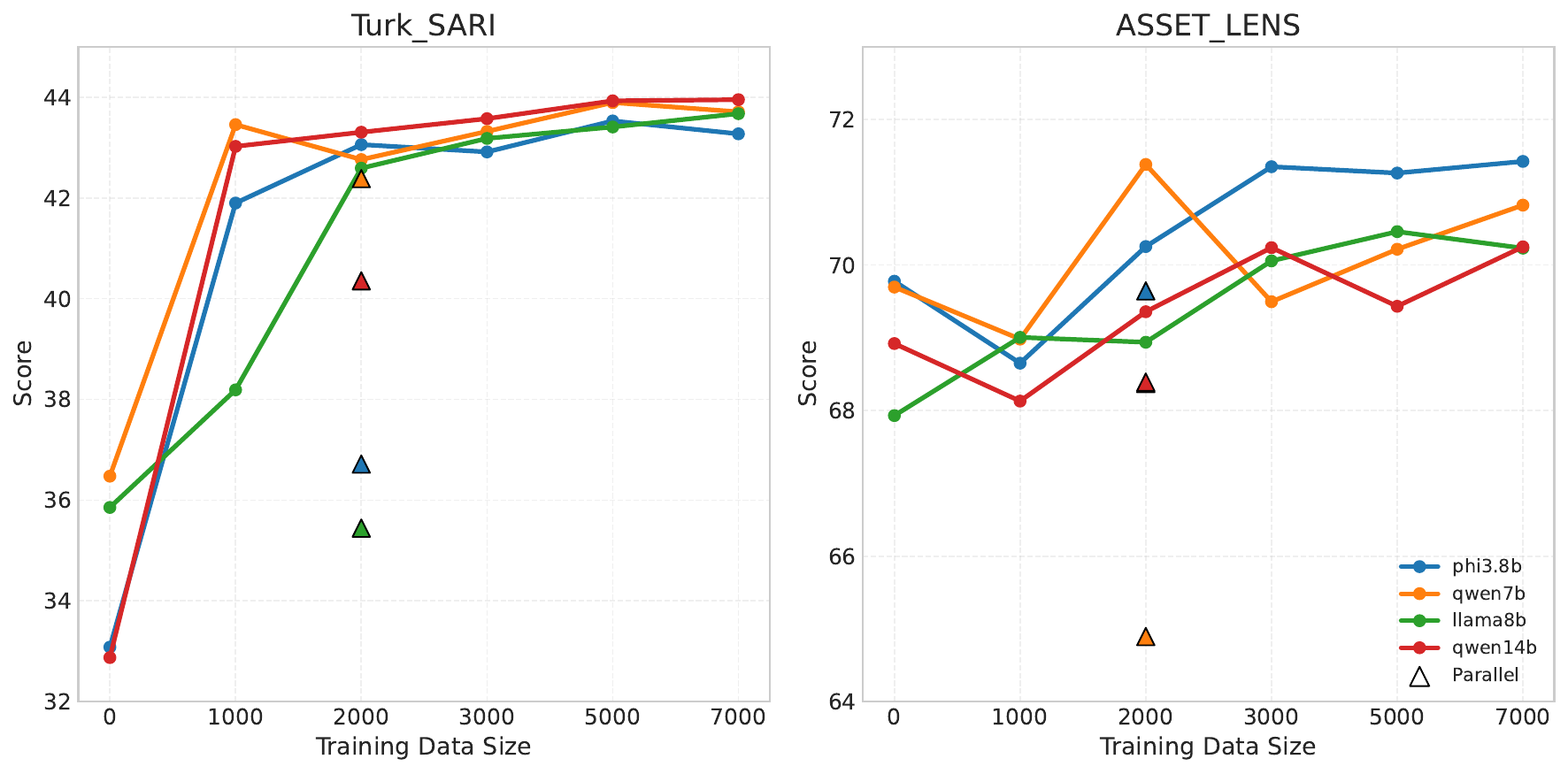}
  \caption{Impact of training sample size in sentence simplification.  Colored lines show models trained on our preference data; colored triangles show models trained on $2k$ human-written parallel data.  Overlap in triangles is due to the nearly identical LENS scores from Qwen14B and Llama8B.}
  \label{fig:train_size_sent}
\end{figure*}

\begin{figure*}[t]
  \small
  \centering
  \includegraphics[width=0.85\linewidth]{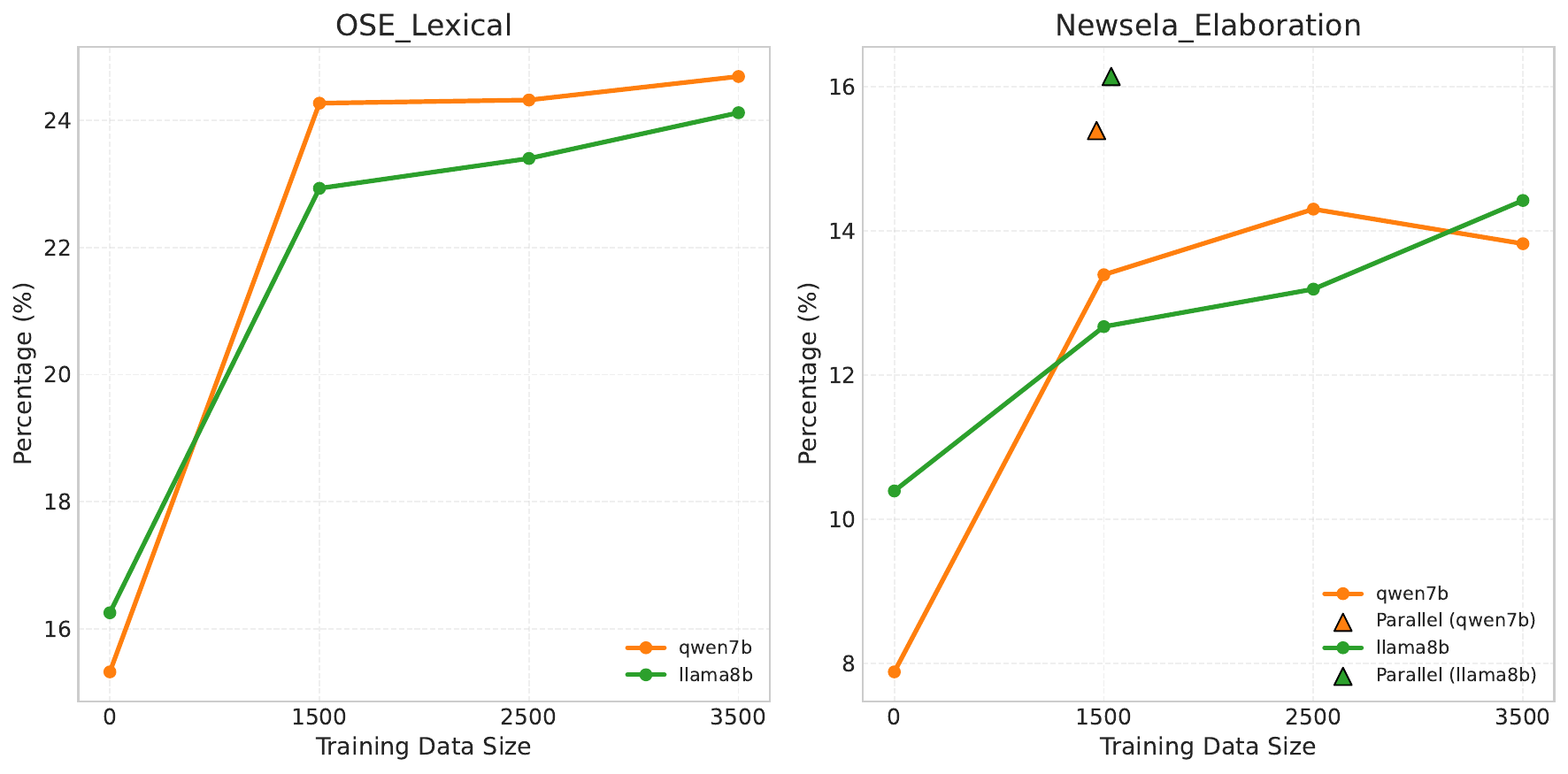}
  \caption{Impact of training sample size in document simplification.}
  \label{fig:train_size_doc}
\end{figure*}

\paragraph{Policy alignment improves as the preference optimization dataset grows.}
Sentence-level results are shown in Figure~\ref{fig:train_size_sent}, and document-level results are shown in Figure~\ref{fig:train_size_doc}. 
The overall trend is clear: performance improves as the training set grows. At the sentence level, models show substantial gains once the training size reaches $2$k, after which improvements become more gradual. One exception is Qwen7B on Sent-Overall, where performance drops when the number of samples increases from $2$k to $3$k, but rises again with further increases in training size.
At the document level, the proportion of desired edits generally increases with training size. Interestingly, for the Doc-Elaboration policy, the two models show different trends from $2.5$k to $3.5$k: Llama increases its elaboration rate, whereas Qwen shows a slight decline. Nevertheless, the Parallel baseline still achieves better performance. These results suggest that learning to perform elaborative edits is more challenging and may require professionally rewritten simplifications, such as those in Newsela, to provide more effective supervision during training.

\subsection{Generalization to Other Domains}
\label{app:ood}
To assess our generalisation capability to other domains, we further evaluated our methods on sentence-level simplification on an out-of-domain of Wikipedia (ASSET and Turk).\footnote{Document-level evaluation was impossible due to the lack of available corpora.}
Specifically, we used \textbf{SimPA}, a public-administration dataset which contains $1, 100$ original sentences paired with two types of references: (1) lexical simplifications, and (2) overall simplifications combined with lexical and syntactic edits. We evaluate our Sent-Lexical model on the former and the Sent-Overall model on the latter. We applied our policy models directly using the same prompts and hyperparameters as in the wiki-domain experiments, without any retraining. Table~\ref{tab:ood_results} shows that our policy models consistently outperform the vanilla baselines. 
It clearly demonstrates that the learned policies are effective across domains.

\begin{table}[t]
    \centering
    \small
    \resizebox{0.85\columnwidth}{!}{ 
    \begin{tabular}{l l c c}
    \toprule
    Model & Method & L-SARI & O-LENS \\
    \midrule
    \multirow{2}{*}{Phi3.8B}
    & Vanilla-zs & $22.3$ & $60.1$ \\
    & PO (Ours) & $\bm{36.9}$ & $\bm{63.5}$ \\
    \midrule
    \multirow{2}{*}{Qwen7B}
    & Vanilla-zs & $26.5$ & $56.0$ \\
    & PO (Ours) & $\bm{36.8}$ & $\bm{63.3}$ \\
    \midrule
    \multirow{2}{*}{Llama8B}
    & Vanilla-zs & $27.7$ & $59.8$ \\
    & PO (Ours) & $\bm{37.1}$ & $\bm{61.2}$ \\
    \midrule
    \multirow{2}{*}{Qwen14B}
    & Vanilla-zs & $27.5$ & $58.7$ \\
    & PO (Ours) & $\bm{37.5}$ & $\bm{61.9}$ \\
    \bottomrule
    \end{tabular}
    }
    \caption{Out-of-domain evaluation results on SimPA. We report results for two policy types: Sent-Lexical (L) and Sent-Overall (O).}
    \label{tab:ood_results}
\end{table}

\subsection{Discussion}
\subsubsection{Qwen3-32B as Candidate Generator, LLM-Judge, and Parser}
\label{sec:circulate_qwen32b}

Qwen3-32B plays three roles in our pipeline: it is used as one of the simplification candidate generators, as the judge model, and as the parser for producing parse-tree information. In this section, we first clarify the motivation for this design and then empirically examine whether its circulation usage introduces systematic bias.

\paragraph{Qwen3-32B is included as a simplification candidate because of its strong simplification performance.}

We use multiple LLMs with different architectures and parameter scales, including Qwen, Phi, and Llama, to increase the diversity of simplification candidates and reduce dependence on a single model family~\footnote{The inclusion of outputs from models that are also used for training, namely Qwen2.5-7B and Llama3.1-8B, is motivated by prior work showing that self-generated outputs can provide useful learning signals for preference optimization~\citep{chen-etal-2024-self-play, wang-etal-2024-self-training}.}. In preliminary experiments, we evaluated several larger open-source LLMs that could be run locally within our computational environment, including Gemma3-27B~\citep{gemmateam2025gemma3technicalreport}, OSS-20B~\citep{openai2025gptoss120bgptoss20bmodel}, and Aya-Expanse-32B~\citep{dang2024ayaexpansecombiningresearch}. Among these models, Qwen3-32B achieved the strongest overall simplification performance. Under the same few-shot prompting setting, Qwen3-32B outperformed the other candidate models on both Turk and ASSET, as shown in Table~\ref{tab:qwen3_candidate_comparison}. We therefore include Qwen3-32B as one of the simplification candidate models, expecting it to enrich the candidate pool with higher-quality simplifications.

\paragraph{Qwen3-32B is selected as the judge due to its strong judge ability.}
The choice of judge model is critical for constructing reliable preference data that aligns with the target edit policies. We compare Qwen3-32B and Gemma3-27B as judge models by using the preference data selected by each model to train downstream models, while keeping all other settings unchanged. As shown in Table~\ref{tab:judge_compare}, models trained on Qwen3-32B-selected preference data consistently outperform those trained on Gemma3-27B-selected data, particularly at the document level. These results suggest that Qwen3-32B provides higher-quality preference annotations for our setting, which in turn improves downstream editing performance.

\begin{table}[t]
\centering
\small
\begin{tabular}{lcc}
\hline
Model & Turk-SARI & ASSET-LENS \\
\hline
Gemma3-27B & $39.5$ & $67.9$ \\
OSS-20B & $41.7$ & $66.3$ \\
Aya-Expanse-32B & $40.4$ & $67.5$ \\
Qwen3-32B & $\bm{43.0}$ & $\bm{71.0}$ \\
\hline
\end{tabular}
\caption{Comparison of larger open-source LLMs under the same few-shot prompting setting.}
\label{tab:qwen3_candidate_comparison}
\end{table}

\begin{table}[t]
\centering
\small

\resizebox{\columnwidth}{!}{
\begin{subtable}{0.95\columnwidth}
\centering
\begin{tabular}{llcc}
\hline
Model & Method & Turk-SARI & ASSET-LENS \\
\hline
Phi3.8B & PO\_Gemma & $42.9$ & $70.5$ \\
        & PO\_Qwen  & $\bm{43.3}$ & $\bm{71.4}$ \\
\hline
Qwen7B  & PO\_Gemma & $43.2$ & $70.2$ \\
        & PO\_Qwen  & $\bm{43.7}$ & $\bm{70.8}$ \\
\hline
Llama8B & PO\_Gemma & $42.9$ & $69.5$ \\
        & PO\_Qwen  & $\bm{43.7}$ & $\bm{70.2}$ \\
\hline
Qwen14B & PO\_Gemma & $43.1$ & $70.0$ \\
        & PO\_Qwen  & $\bm{43.9}$ & $\bm{70.3}$ \\
\hline
\end{tabular}
\caption{Sentence-level}
\end{subtable}
}
\vspace{0.8em}

\resizebox{\columnwidth}{!}{
\begin{subtable}{0.95\columnwidth}
\centering
\resizebox{\columnwidth}{!}{%
\begin{tabular}{llcccc}
\hline
& & \multicolumn{2}{c}{OSE} & \multicolumn{2}{c}{Newsela} \\
\cline{3-6}
Model & Method & LENS & BERT & LENS & BERT \\
\hline
Qwen7B & PO\_Gemma & $50.7$ & $45.6$ & $49.1$ & $49.0$ \\
       & PO\_Qwen  & $\bm{53.7}$ & $\bm{50.1}$ & $\bm{50.7}$ & $\bm{52.0}$ \\
\hline
Llama8B & PO\_Gemma & $51.9$ & $47.7$ & $51.1$ & $52.4$ \\
        & PO\_Qwen  & $\bm{53.9}$ & $\bm{51.5}$ & $\bm{52.6}$ & $\bm{52.6}$ \\
\hline
\end{tabular}%
}
\caption{Document-level}
\end{subtable}
}
\caption{Comparison between Gemma3-27B and Qwen3-32B as the judge.}
\label{tab:judge_compare}
\end{table}

\begin{table*}[t]
\centering
\small
\resizebox{\textwidth}{!}{
\begin{tabular}{lcccccc}
\hline
Model
& \multicolumn{4}{c}{Sentence-level}
& \multicolumn{2}{c}{Document-level} \\
\cline{2-5} \cline{6-7}
& \multicolumn{2}{c}{w/o Parse}
& \multicolumn{2}{c}{w/ Parse}
& Doc-Lexical
& Doc-Elaboration \\
\cline{2-3} \cline{4-5}
& Sent-Lexical
& Sent-Overall
& Sent-Lexical
& Sent-Overall
& & \\
\hline
Qwen2.5-7B
& $24.0$/$22.3$ & $25.0$/$22.9$
& $22.9$/$22.3$ & $25.1$/$22.2$
& $19.5$/$17.3$ & $21.1$/$15.5$ \\
Llama3.1-8B
& $12.1$/$50.5$ & $14.3$/$48.7$
& $13.2$/$49.9$ & $13.7$/$48.8$
& $5.9$/$63.1$ & $6.4$/$68.3$ \\
Phi4-14B
& $31.6$/$12.1$ & $31.2$/$11.3$
& $32.0$/$12.3$ & $32.5$/$11.4$
& $30.3$/$12.3$ & $30.5$/$10.2$ \\
Qwen3-32B
& $32.2$/$15.0$ & $29.6$/$17.2$
& $31.9$/$15.5$ & $28.8$/$17.6$
& $44.2$/$7.2$ & $42.0$/$6.0$ \\
\hline
\end{tabular}
}
\caption{Preference distribution under the Qwen3-32B judge. Each value is reported as preferred/dispreferred ratio. For sentence-level criteria, we also report ablation results without parse-tree information.}
\label{tab:qwen_bias_analysis}
\end{table*}

\paragraph{Qwen-generated parse trees do not bias the judge toward Qwen3-32B outputs.}
We examine whether using Qwen3-32B to generate parse-tree information biases the judge toward Qwen3-32B outputs. We conduct an ablation study in which only the parse-tree information is removed, while all other components remain unchanged. If Qwen-generated parse trees induced a preference for Qwen3-32B outputs, we would expect the preference fraction of Qwen3-32B to increase when parse-tree information is included.
However, we observe the opposite pattern as shown in Table~\ref{tab:qwen_bias_analysis}. Adding parse-tree information slightly decreases the preferred fraction of Qwen3-32B in both Sent-Lexical and Sent-Overall settings: from $32.2\%$ to $31.9\%$ and from $29.6\%$ to $28.8\%$, respectively. This indicates that the use of Qwen-generated parse trees does not lead the judge to favor Qwen3-32B outputs.

\paragraph{Qwen3-32B as the judge does not systematically favor its own simplification outputs.}
We examine whether using Qwen3-32B as both the judge and one of the candidate generators introduces model-selection bias. If the judge systematically favored its own generations, we would expect outputs generated by Qwen3-32B to dominate the selected candidates across edit policies. To test this, we analyze the model-wise distribution of preferred candidates selected by Qwen3-32B. As shown in Table~\ref{tab:qwen_bias_analysis}, the preferred fraction of Qwen3-32B outputs ranges from $28.8\%$ to $44.2\%$ across policies, and Qwen3-32B is not consistently the most frequently selected candidate model. For example, Phi4-14B is selected more often than Qwen3-32B under both sentence-level policies and also receives high preferred fractions in the document-level settings. These results suggest that Qwen3-32B, when used as the judge, does not simply select its own outputs by default.

\subsubsection{Model Outputs vs. Human}
To better understand the discrepancies between edit patterns in PO model outputs and human references discussed in Section~\ref{sec:edit_pattern}, we manually examined a small set of examples. The case analysis shows that human references and PO model outputs often differ in both what they simplify and how they simplify it. For example, human references sometimes simplify domain-specific terms, while PO models tend to preserve them. PO models also often paraphrase the surrounding context more broadly than human references. An illustrative example from the Doc-Lexical policy is shown below.

\begin{description}
    \item[Source:] ...the 1961 Single Convention on Narcotic Drugs, the framework that governs international drugs policy...
    \item[Human:] ...the 1961 Convention on Drugs, the agreement that controls international drugs policy...
    \item[Qwen7B:] ...the 1961 Single Convention on Narcotic Drugs, which sets rules for international drug policy...
    \item[Llama8B:] ...the 1961 Single Convention on Narcotic Drugs, which is the main rule for international drug policy...
\end{description}

The human reference simplifies the domain-specific term \textit{1961 Single Convention on Narcotic Drugs} to \textit{1961 Convention on Drugs}, while both PO models preserve the full term. The human reference also makes localized lexical substitutions, such as \textit{framework} $\rightarrow$ \textit{agreement} and \textit{governs} $\rightarrow$ \textit{controls}. In contrast, the PO models paraphrase the surrounding phrase more broadly, using expressions such as \textit{sets rules for} and \textit{is the main rule for}. These differences can reduce measured span overlap, even when the model outputs constitute valid simplifications.

\begin{table*}[t]
\centering
\small
\setlength{\tabcolsep}{4pt}
\renewcommand{\arraystretch}{1.15}
\begin{tabularx}{\textwidth}{p{0.15\textwidth} X}
\toprule
\textbf{Model} & \textbf{Sentence} \\
\midrule

Source 1 & One side of the armed conflicts is composed mainly of the Sudanese military and the Janjaweed, a Sudanese militia group recruited mostly from the Afro-Arab Abbala tribes of the northern Rizeigat region in Sudan. \\
\hdashline
Phi3B-Lexical & One side of the armed conflicts is mainly made up of the Sudanese military and the Janjaweed, a Sudanese militia group mostly recruited from the Afro-Arab Abbala tribes in the northern Rizeigat region of Sudan. \\
Phi3B-Overall & The armed conflicts involve the Sudanese military and the Janjaweed, a Sudanese militia group mostly made up of the Afro-Arab Abbala tribes from the northern Rizeigat region in Sudan. \\
\hdashline
Qwen7B-Lexical & One side of the armed conflicts is mainly made up of the Sudanese military and the Janjaweed, a Sudanese militia group mostly from the Afro-Arab Abbala tribes in the northern Rizeigat region of Sudan. \\
Qwen7B-Overall & One side of the armed conflicts includes the Sudanese military and the Janjaweed, a militia group from the northern Rizeigat region. \\
\hdashline
Llama8B-Lexical & One side of the conflicts is made up mainly of the Sudanese military and the Janjaweed, a militia group from the Afro-Arab Abbala tribes in northern Sudan. \\
Llama8B-Overall & One side of the armed conflicts includes the Sudanese military and the Janjaweed, a militia group from the northern Rizeigat region, mostly from the Afro-Arab Abbala tribes. \\
\hdashline
Qwen14B-Lexical & One side of the conflict is made up mainly of the Sudanese military and the Janjaweed, a Sudanese militia group mostly recruited from the Afro-Arab Abbala tribes in the northern Rizeigat region of Sudan. \\
Qwen14B-Overall & One side of the armed conflicts includes the Sudanese military and the Janjaweed, a Sudanese militia group from the northern Rizeigat region. \\

\midrule

Source 2 & The spacecraft consists of two main elements: the NASA Cassini orbiter, named after the Italian-French astronomer Giovanni Domenico Cassini, and the ESA Huygens probe, named after the Dutch astronomer, mathematician and physicist Christiaan Huygens. \\
\hdashline
Phi3B-Lexical & The spacecraft has two main parts: the NASA Cassini orbiter, named after Italian-French astronomer Giovanni Domenico Cassini, and the ESA Huygens probe, named after Dutch astronomer, mathematician, and physicist Christiaan Huygens. \\
Phi3B-Overall & The spacecraft has two main parts: the NASA Cassini orbiter and the ESA Huygens probe. \\
\hdashline
Qwen7B-Lexical & The spacecraft has two main parts: the NASA Cassini orbiter, named after the Italian-French astronomer Giovanni Domenico Cassini, and the ESA Huygens probe, named after the Dutch astronomer, mathematician, and physicist Christiaan Huygens. \\
Qwen7B-Overall & The spacecraft has two main parts: the NASA Cassini orbiter, named after Giovanni Domenico Cassini, and the ESA Huygens probe, named after Christiaan Huygens. \\
\hdashline
Llama8B-Lexical & The spacecraft has two main parts: the NASA Cassini orbiter, named after Italian-French astronomer Giovanni Domenico Cassini, and the ESA Huygens probe, named after Dutch astronomer, mathematician, and physicist Christiaan Huygens. \\
Llama8B-Overall & The spacecraft has two main parts: the NASA Cassini orbiter, named after Giovanni Domenico Cassini, and the ESA Huygens probe, named after Christiaan Huygens. \\
\hdashline
Qwen14B-Lexical & The spacecraft has two main parts: the NASA Cassini orbiter, named after the Italian-French astronomer Giovanni Domenico Cassini, and the ESA Huygens probe, named after the Dutch astronomer, mathematician, and physicist Christiaan Huygens. \\
Qwen14B-Overall & The spacecraft has two main parts: the NASA Cassini orbiter, named after Giovanni Domenico Cassini, and the ESA Huygens probe, named after Christiaan Huygens. \\

\midrule

Source 3 & Each entry has a datum (a nugget of data) which is a copy of the datum in some backing store. \\
\hdashline
Phi3.8B-Lexical & Each entry has a piece of data (a nugget of data) that is a copy of the data in some storage. \\
Phi3.8B-Overall & Each entry has a piece of data that is a copy from a storage. \\
\hdashline
Qwen7B-Lexical & Each entry has a datum, which is a copy of the data from some storage. \\
Qwen7B-Overall & Each entry has a copy of the data from a backing store. \\
\hdashline
Llama8B-Lexical & Each entry has a copy of the data from the backing store. \\
Llama8B-Overall & Each entry has a copy of the data from a backing store. \\
\hdashline
Qwen14B-Lexical & Each entry has a piece of data that is a copy of the data in a backup storage. \\
Qwen14B-Overall & Each entry is a copy of a datum from a backing store. \\

\bottomrule
\end{tabularx}
\caption{Examples of Sent-Lexical and Sent-Overall simplifications by our models.}
\label{tab:simp_examplessent}
\end{table*}

\begin{table*}[t]
\centering
\small
\setlength{\tabcolsep}{4pt}
\renewcommand{\arraystretch}{1.15}
\begin{tabularx}{\textwidth}{p{0.1\textwidth} X}
\toprule
\textbf{Model} & \textbf{Document} \\
\midrule

Source 1 & Dame Sally Davies, the Chief Medical Officer for England, has likened the problem of antibiotic resistance to the risks presented by international terrorism. While this might sound like an exaggeration, the threat was actually, if anything, understated. Each year, the global number of deaths to which bacterial resistance contributes far outstrips those caused by terrorist attacks. While it is difficult to track the global impact of antibiotic resistance across all bacterial species, the World Health Organization estimates that for tuberculosis alone multi-drug resistance accounts for more than 150,000 deaths each year. Antibiotic resistance is no longer an abstract risk: this is now a war. In the past hundred years, our expectations of life and survival have changed beyond all recognition... \\
\hdashline
Qwen7B-Lexical & Dame Sally Davies, the Chief Medical Officer for England, has compared the threat of antibiotic resistance to international terrorism. This comparison is not an exaggeration; the problem is even more serious than it might seem. Each year, the number of deaths caused by antibiotic-resistant bacteria is much higher than those caused by terrorist attacks. While it's hard to track the global impact of antibiotic resistance, the World Health Organization estimates that multi-drug resistant tuberculosis alone causes more than 150,000 deaths annually. Antibiotic resistance is no longer just a potential risk; it's a real and urgent threat. In the past century, our expectations of life and survival have changed dramatically... \\
\hdashline
Llama8B-Lexical & Dame Sally Davies, the Chief Medical Officer for England, has compared the problem of antibiotic resistance to the risks of international terrorism. While this might seem like an exaggeration, the threat is actually more serious. Each year, the number of deaths caused by bacterial resistance is much higher than those caused by terrorist attacks. The World Health Organization estimates that for tuberculosis alone, multi-drug resistance causes more than 150,000 deaths each year. Antibiotic resistance is no longer just a risk; it's a real war. In the past hundred years, our expectations of life and survival have changed dramatically... \\

\midrule

Source 2 &  Standing at the edge of space above the deserts of New Mexico, Felix Baumgartner paused slightly. It was a small step away from the capsule, but a 24-mile drop back down to Earth. Our guardian angel will take care of you, said mission control, and the man known as Fearless Felix jumped. Ten heart-stopping minutes later the Austrian landed back on Earth, after reaching speeds of up to 725mph, and breaking three world records, including becoming the worlds first supersonic skydiver by breaking the sound barrier at Mach 1.24. We love you Felix, cheered the control room as his mother, Ava Baumgartner, wept. Baumgartner, who claimed the records for the highest altitude manned balloon flight and the highest altitude skydive, raised his arms in a victory salute to thank his team... \\
\hdashline
Qwen7B-Lexical & Felix Baumgartner stood at the edge of space above the deserts of New Mexico. He was about to jump from a capsule 24 miles above the Earth. Mission control said, "Our guardian angel will take care of you." Baumgartner jumped. Ten minutes later, he landed safely on Earth, having reached speeds of up to 725 mph and breaking three world records, including becoming the first supersonic skydiver by breaking the sound barrier at Mach 1.24. The control room cheered, and Baumgartner's mother, Ava Baumgartner, cried. He raised his arms in a victory salute to thank his team... \\
\hdashline
Llama8B-Lexical & Felix Baumgartner, known as Fearless Felix, stood at the edge of space above the deserts of New Mexico. He was about to jump from a capsule 24 miles above the Earth. Mission control told him, "Our guardian angel will take care of you." Felix jumped, and 10 minutes later, he landed back on Earth, reaching speeds of up to 725mph. He broke three world records, including becoming the first person to break the sound barrier in free-fall. The control room cheered, and his mother, Ava Baumgartner, cried. Felix raised his arms in a victory salute to thank his team... \\

\midrule
Source 3 &  A major international row with wide-ranging 1 implications for global drugs policy has erupted over the right of Bolivias indigenous Indian tribes to chew coca leaves, the principal ingredient in cocaine. Bolivia has obtained a special exemption from the 1961 Single Convention on Narcotic Drugs, the framework that governs international drugs policy, allowing its indigenous people to chew the leaves. Bolivia had argued that the convention was 2 in opposition to its new constitution, adopted in 2009, which obliges it to protect native and ancestral coca as cultural patrimony and maintains that coca in its natural state is not a narcotic. South American Indians have chewed coca 3 leaves for centuries... \\
\hdashline
Qwen7B-Lexical &  A big international argument has started over the right of Bolivia's indigenous people to chew coca leaves, which are used to make cocaine. Bolivia has asked to be allowed to keep this tradition because it is part of their culture. The 1961 Single Convention on Narcotic Drugs, which sets rules for international drug policy, usually doesn't allow this. But Bolivia got a special permission to let its indigenous people chew the leaves. This is because their new constitution, from 2009, says they should protect coca as part of their culture and says coca in its natural form is not a drug.South American Indians have chewed coca leaves for centuries... \\
\hdashline
Llama8B-Lexical & A big international argument has started about the right of Bolivia's indigenous people to chew coca leaves, which are the main ingredient in cocaine. Bolivia has asked for a special exception from the 1961 Single Convention on Narcotic Drugs, which is the main rule for international drug policy. This rule says that coca leaves are illegal, but Bolivia says that in its natural state, they are not a drug and are important to its culture. The country's new constitution, adopted in 2009, says that coca is a part of its cultural heritage and should be protected. South American Indians have been chewing coca leaves for centuries, and they are... \\

\bottomrule
\end{tabularx}
\caption{Examples of Doc-Lexical simplifications by our models. Documents are truncated for readability.}
\label{tab:simp_examplesdoc}
\end{table*}

\begin{table*}[t]
    \centering
    \small
    \setlength{\tabcolsep}{5pt}
    \renewcommand{\arraystretch}{1.12}
    \begin{tabular}{@{}p{0.14\linewidth} p{0.18\linewidth} p{0.62\linewidth}@{}}
    \toprule
    {} & {} & \textbf{Content} \\
    \midrule
    
    \textbf{Source} & \multicolumn{2}{p{0.80\linewidth}@{}}{%
    Unveiling a car with a top speed of 25mph, two seats, and no pedals or steering wheel might not make much of an impression at a motor show. But Google, in the US, sent a minor earthquake through the car and taxi industries as it unveiled the latest version of its driverless car. The electrically powered vehicle, which Google has begun testing around its headquarters in Mountain View, California, dispenses with all the normal controls, including foot pedals. Instead, it has a smartphone app that summons it and tells it the destination, and a single STOP button mounted between the two front-facing seats in case the occupants need to override the computer. \textit{[truncated]}} \\
    \midrule
    
    \textbf{Simplified} & \multicolumn{2}{p{0.80\linewidth}@{}}{%
    A car with a maximum speed of 25 miles per hour, two seats, and no pedals or steering wheel does not sound very interesting. But Google, in the US, shocked the car and taxi industries when it unveiled the latest version of its driverless car. Google has begun testing the electric car at its headquarters in Mountain View, California. The car does not have the normal controls, such as foot pedals. Instead, it has a smartphone app that calls it and tells it the destination, and a STOP button between the two seats in case the passengers need to override the computer. \textit{[truncated]}} \\
    \midrule
    
    {\textbf{Edit Category}}
    & \textbf{Lexical Generic}
    & \textit{Location 1} \\
    && ``source'': ``Unveiling a car with a top \ldots\ at a motor show.'' \\
    && ``simplified'': ``A car with a maximum \ldots\ very interesting.'' \\
    && ``edit'': ``A'' $\leftarrow$ ``Unveiling a''; ``maximum'' $\leftarrow$ ``top''; \ldots\ \textit{[truncated]} \\
    && \ldots\ \textit{[truncated]} \\
    && \textit{Location 21} \\
    && ``source'': ``The company points out that \ldots\ due to human error.'' \\
    && ``simplified'': ``The company says that \ldots\ by human mistakes.'' \\
    && ``edit'': ``says'' $\leftarrow$ ``points out''; ``mistakes'' $\leftarrow$ ``error'' \\
    \cmidrule(l){2-3}
    
    & \textbf{Nonsim Format}
    & \textit{Location 1} \\
    && ``source'': ``Unveiling a car with \ldots\ at a motor show.'' \\
    && ``simplified'': ``A car with a maximum \ldots\ very interesting.'' \\
    && ``edit'': ``25 miles per hour'' $\leftarrow$ ``25mph''; \ldots\ \textit{[truncated]} \\
    \cmidrule(l){2-3}
    
    & \textbf{Discourse Reordering}
    & \textit{Location 1} \\
    && ``source'': ``The electrically powered vehicle, which Google has begun testing around its headquarters in Mountain View, California, dispenses with all the normal controls, including foot pedals.'' \\
    && ``simplified'': ``Google has begun testing the electric car at its headquarters in Mountain View, California.'' \\
    && ``edit'': ``The electrically powered vehicle, which''; \ldots\ \textit{[truncated]} \\
    \cmidrule(l){2-3}
    
    & \textit{\ldots\ \textit{[truncated]}}
    & \textit{...} \\
    \bottomrule
    \end{tabular}
    \caption{Anchor example for Doc-Lexical policy used in the prompt for LLM-Judge. This example is a held-out sample from the OSE dataset. For readability, we truncate each field because the documents and edit annotations are very long.}
    \label{tab:anchor_ose}
\end{table*}


\begin{figure*}[t]
  \centering
  \begin{subfigure}{\linewidth}
    \centering
    \includegraphics[width=0.8\linewidth]{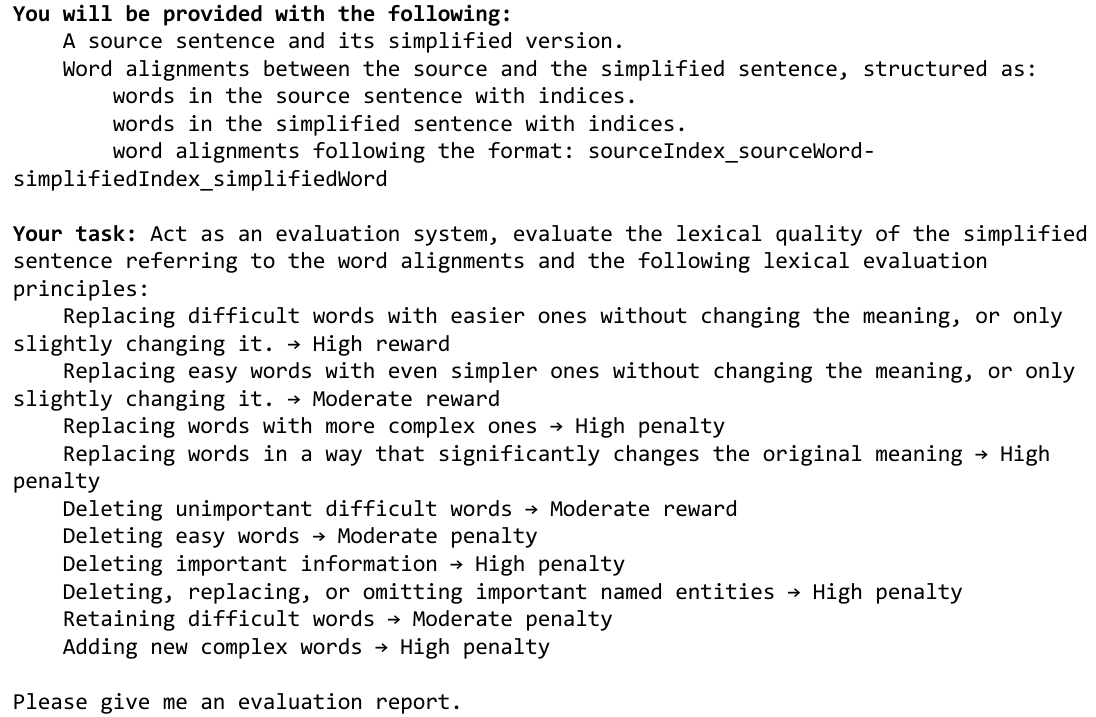}
    \caption{Sent-Lexical}
  \end{subfigure}
  \begin{subfigure}{\linewidth}
    \centering
    \includegraphics[width=0.8\linewidth]{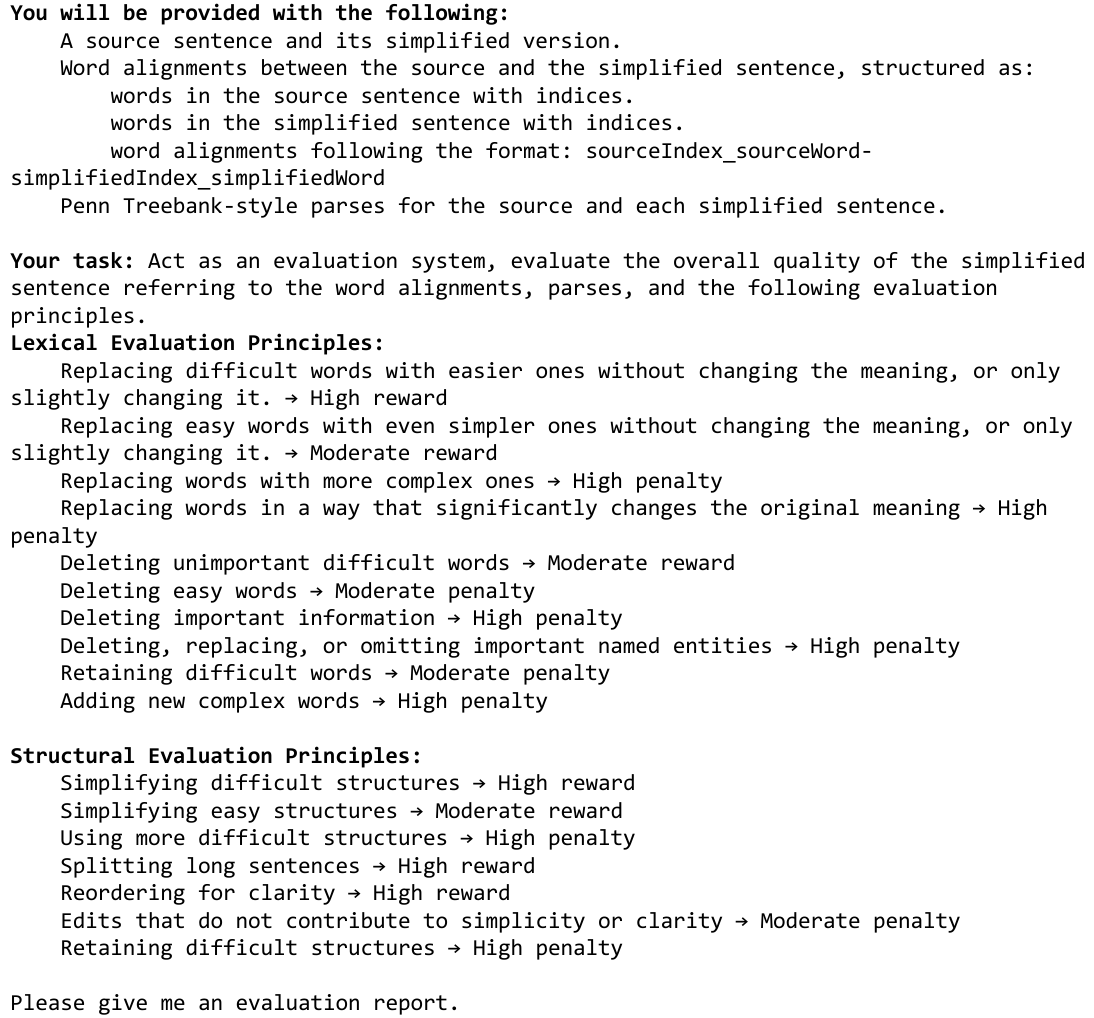}
    \caption{Sent-Overall}
  \end{subfigure}
  \caption{Guidelines in sentence-level iterative feedback. Contents are from Figure~\ref{fig:guidlines_sent}, ~\ref{fig:judge_example1_sent}, ~\ref{fig:judge_example2_sent}, and ~\ref{fig:judge_example3_sent}.}
  \label{fig:prompt_guide_iter_sent}
\end{figure*}

\begin{figure*}[t]
  \centering
  \begin{subfigure}{\linewidth}
    \centering
    \includegraphics[width=0.8\linewidth]{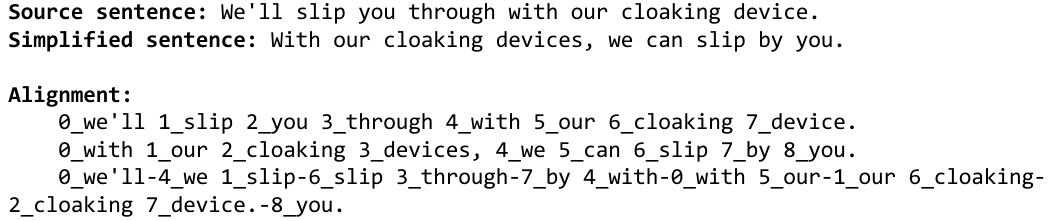}
    \caption{Input of Example 1}
  \end{subfigure}
  \begin{subfigure}{\linewidth}
    \centering
    \includegraphics[width=0.8\linewidth]{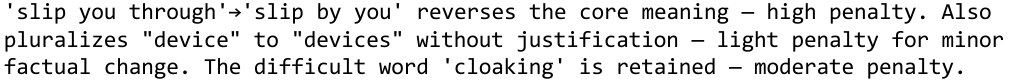}
    \caption{Output of Example 1}
  \end{subfigure}
  \begin{subfigure}{\linewidth}
    \centering
    \includegraphics[width=0.8\linewidth]{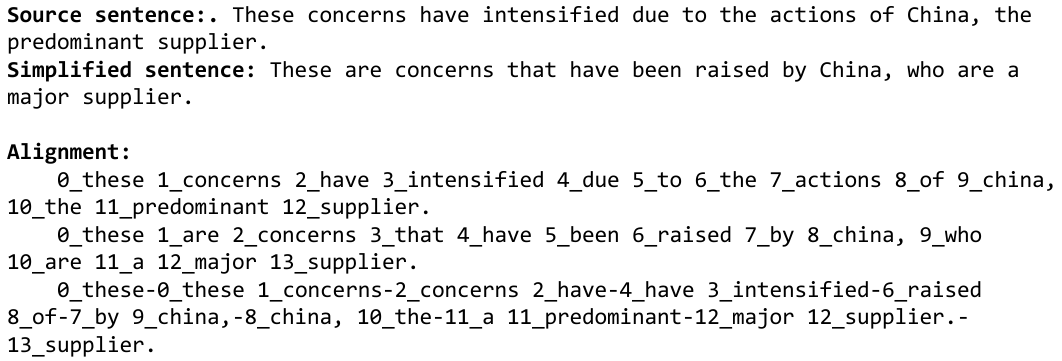}
    \caption{Input of Example 2}
  \end{subfigure}
  \begin{subfigure}{\linewidth}
    \centering
    \includegraphics[width=0.8\linewidth]{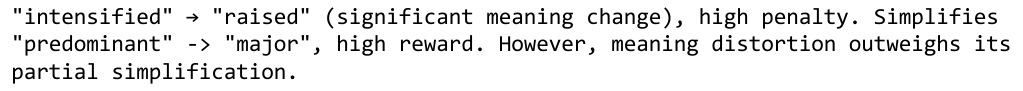}
    \caption{Output of Example 2}
  \end{subfigure}
  \begin{subfigure}{\linewidth}
    \centering
    \includegraphics[width=0.8\linewidth]{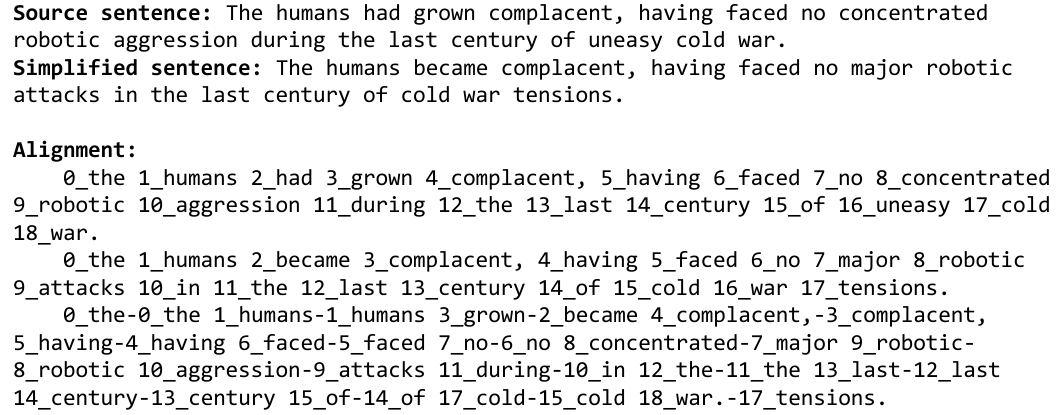}
    \caption{Input of Example 3}
  \end{subfigure}
  \begin{subfigure}{\linewidth}
    \centering
    \includegraphics[width=0.8\linewidth]{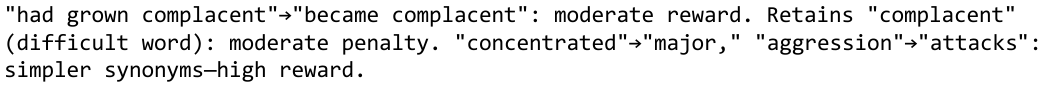}
    \caption{Output of Example 3}
  \end{subfigure}
  \caption{Three-shot examples of Sent-Lexical used for LLM-Judge in the iterative feedback setting.}
  \label{fig:judge_example_iter_lexical_sent}
\end{figure*}

\begin{figure*}[t]
  \centering
  \begin{subfigure}{\linewidth}
    \centering
    \includegraphics[width=0.75\linewidth]{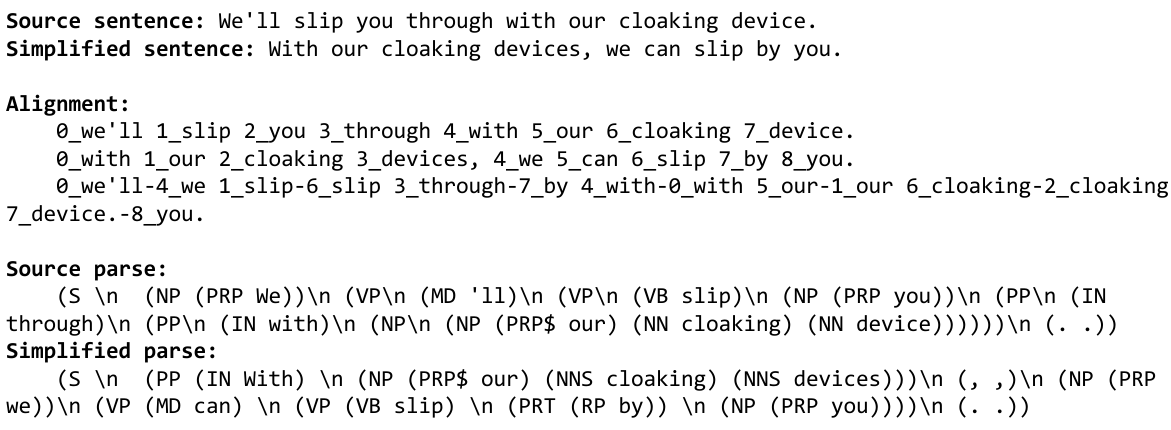}
    \caption{Input of Example 1}
  \end{subfigure}
  \begin{subfigure}{\linewidth}
    \centering
    \includegraphics[width=0.75\linewidth]{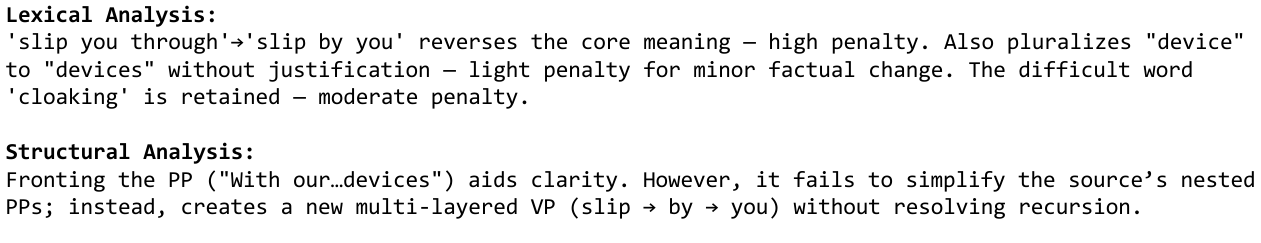}
    \caption{Output of Example 1}
  \end{subfigure}
  \begin{subfigure}{\linewidth}
    \centering
    \includegraphics[width=0.75\linewidth]{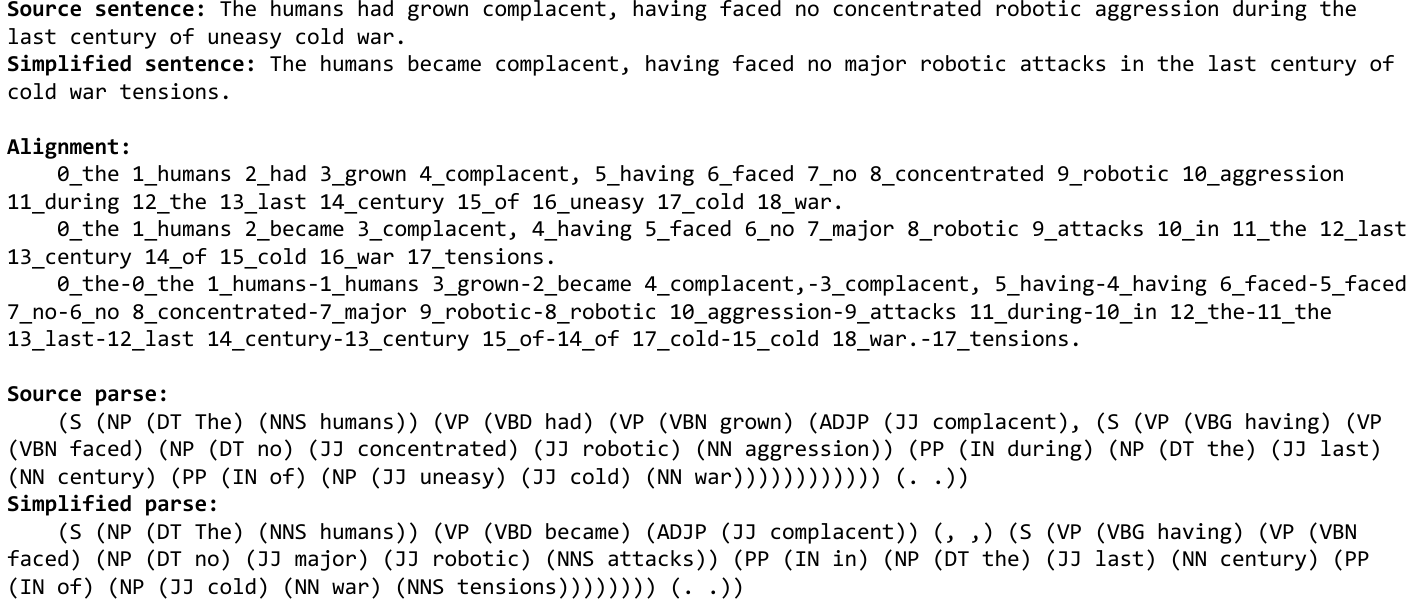}
    \caption{Input of Example 2}
  \end{subfigure}
  \begin{subfigure}{\linewidth}
    \centering
    \includegraphics[width=0.75\linewidth]{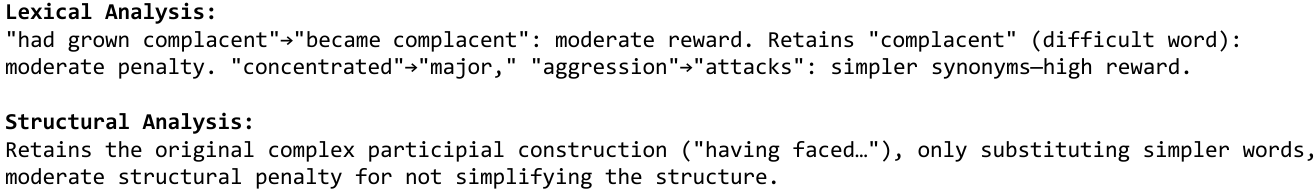}
    \caption{Output of Example 2}
  \end{subfigure}
  \begin{subfigure}{\linewidth}
    \centering
    \includegraphics[width=0.75\linewidth]{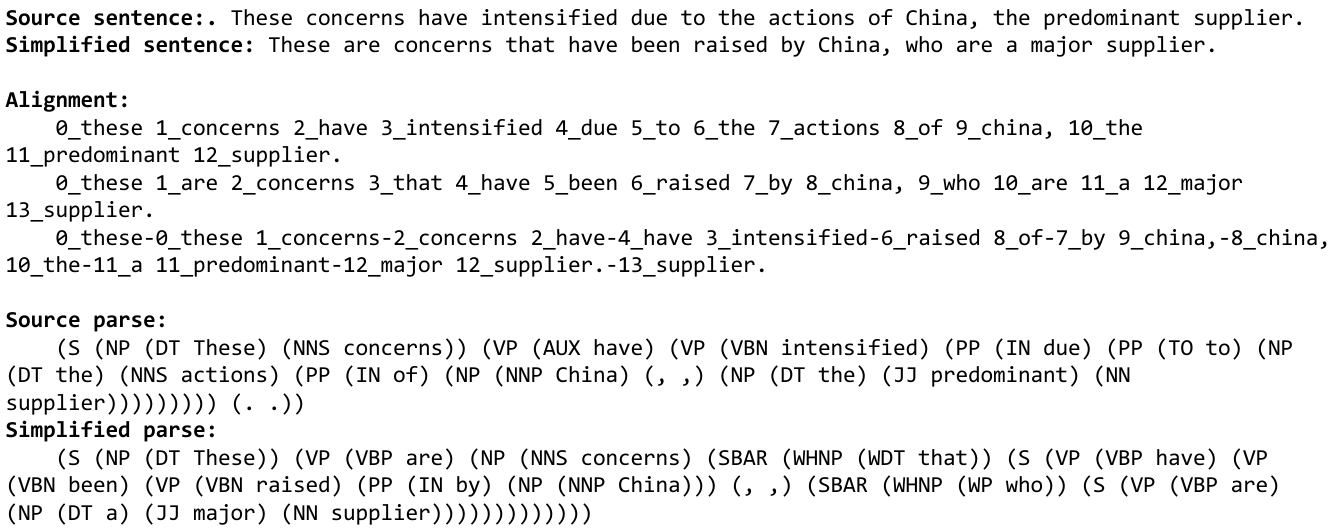}
    \caption{Input of Example 3}
  \end{subfigure}
  \begin{subfigure}{\linewidth}
    \centering
    \includegraphics[width=0.75\linewidth]{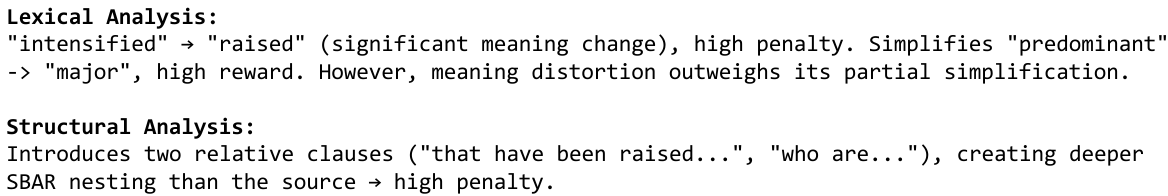}
    \caption{Output of Example 3}
  \end{subfigure}
  \caption{3-shot examples of Sent-Overall used for LLM-Judge in the iterative feedback setting}
  \label{fig:judge_examples_iter_overall_sent}
\end{figure*}

\begin{figure*}[t]
  \centering
  \includegraphics[width=0.95\linewidth]{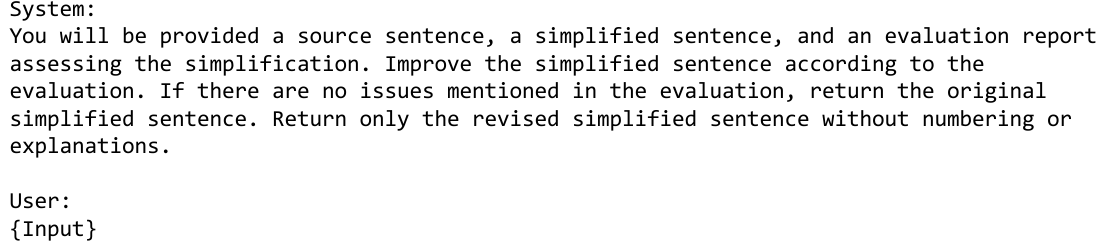}
  \caption{Prompts used for self-correction in the iterative feedback setting at the sentence level}
  \label{fig:correct_sent}
\end{figure*}

\begin{figure*}[t]
  \centering
  \includegraphics[width=0.8\linewidth]{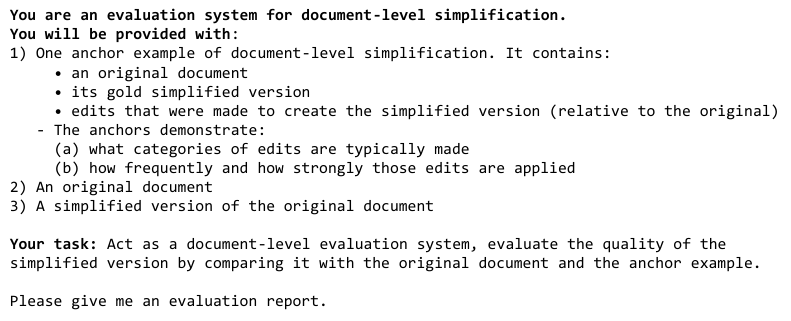}
  \caption{Guidelines in document-level iterative feedback.}
  \label{fig:prompt_guide_iter_doc}
\end{figure*}

\begin{figure*}[t]
  \centering
  \includegraphics[width=0.8\linewidth]{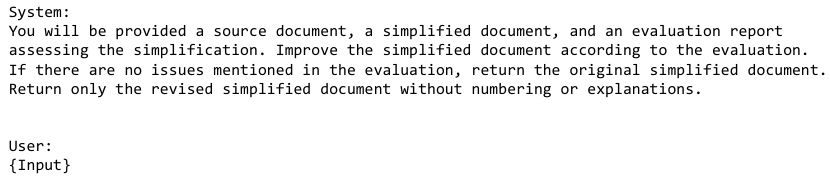}
  \caption{Prompts used for self-correction in the iterative feedback setting at the document level}
  \label{fig:correct_doc}
\end{figure*}

\end{document}